\documentclass{article}

 \usepackage[preprint, nonatbib]{neurips_2026}

\usepackage[utf8]{inputenc} 
\usepackage[T1]{fontenc}    
\usepackage{hyperref}       
\usepackage{url}            
\usepackage{booktabs}       
\usepackage{amsfonts}       
\usepackage{nicefrac}       
\usepackage{microtype}      
\usepackage{xcolor}         
\usepackage{cite}
\usepackage{soul}
\usepackage{subcaption}
\usepackage{wrapfig}

\usepackage{algorithm}
\usepackage{algpseudocode}

\usepackage{amsmath}
\usepackage{amssymb}
\usepackage{mathtools}
\usepackage{amsthm}

\usepackage[capitalize,noabbrev]{cleveref}

\theoremstyle{plain}
\newtheorem{theorem}{Theorem}[section]

\newtheorem{lemma}[theorem]{Lemma}
\newtheorem{corollary}[theorem]{Corollary}
\theoremstyle{definition}
\newtheorem{definition}[theorem]{Definition}
\newtheorem{assumption}[theorem]{Assumption}
\theoremstyle{remark}

\usepackage{amsmath,amsfonts,bm}
\usepackage{enumitem}

\def\eqref#1{equation~\ref{#1}}

\def\1{\bm{1}}

\DeclareMathAlphabet{\mathsfit}{\encodingdefault}{\sfdefault}{m}{sl}
\SetMathAlphabet{\mathsfit}{bold}{\encodingdefault}{\sfdefault}{bx}{n}

\DeclareMathOperator*{\argmin}{arg\,min}

\newcommand{\x}{\mathbf{x}}
\newcommand{\y}{\mathbf{y}}
\newcommand{\z}{\mathbf{z}}
\newcommand{\e}{\mathbf{e}}

\newcommand{\algcd}{$\mathsf{LeGo\text{-}BiO}$}

\title{Lower-Level Agnostic Bilevel Optimization}

\author{%
  Peiwen Qiu \\
  Dept. of ECE\\
  The Ohio State University\\
  Columbus, OH 43210 \\
  \texttt{qiu.617@osu.edu} \\
  \And
  Prashant Khanduri \\
  Dept. of CS \\
  Wayne State University\\
  Detroit, MI 48202 \\
  \texttt{khanduri.prashant@wayne.edu} \\
  \And
  Jia Liu \\
  Dept. of ECE \\
  The Ohio State University\\
  Columbus, OH 43210 \\
  \texttt{liu@ece.osu.edu} \\
}

\begin{document}

\maketitle

\allowdisplaybreaks


\begin{abstract}
Bilevel optimization is a fundamental framework for machine learning problems, where an upper-level (UL) objective depends on the solution of a nested lower-level (LL) problem.
Existing bilevel algorithms typically assume that the LL objective is available for evaluation, differentiation, or unrolling.
However, this assumption can fail when the LL response is produced by a black-box solver, simulator, adversary, or learned predictor.
In this paper, we study \emph{LL agnostic} bilevel optimization, where the UL learner has access only to an approximate LL solution $\bar{\mathbf{y}}(\mathbf{x})$ and cannot query the LL objective, its gradients, Hessian/Jacobian information, or the procedure that generates the LL response.
We propose \algcd~(\ul{L}ower-l\ul{e}vel A\ul{g}n\ul{o}stic \ul{Bi}level \ul{O}ptimization), a coordinate-wise pseudo-hypergradient method that estimates the missing Jacobian $\nabla _{\mathbf{x}}\bar{\mathbf{y}}(\mathbf{x})$ using finite differences of $\bar{\mathbf{y}}(\cdot)$.
For each UL coordinate, \algcd~constructs a decision vector subject to 1-sparse updates, i.e., restricting modifications to a single coordinate per step, thereby isolating the corresponding Jacobian column needed in the UL gradient computation.
This avoids the directional-projection limitation of full-parameter response differences and, unlike derivative-free/zeroth-order approximations of the entire UL gradient, preserves the available first-order gradient information with respect to the UL variables for more accurate UL gradient evaluation.
We establish convergence guarantees for \algcd~in both deterministic and stochastic settings under an LL error-bound/Polyak–Łojasiewicz-type condition, without requiring LL strong convexity.
Our bounds yield an $\mathcal{O}(T^{-1})$ rate in the deterministic setting and an $\mathcal{O}(T^{-1/2})$ rate in the stochastic setting when the LL approximation error decreases geometrically.
Experiments on deep hyper-representation, adversarial training, and data weighting for LLM alignment demonstrate competitive performance of \algcd~despite being agnostic to the LL objective.
\end{abstract}




\section{Introduction} \label{sec: Introduction}

Bilevel optimization provides a natural framework for learning problems in which an upper-level (UL) decision depends on the solution of a nested lower-level (LL) problem.  
A standard bilevel optimization problem is formulated as
\begin{gather} \label{eq: BLO}
\min _{\x \in \mathbb{R}^{d_u}} \ell(\x) \triangleq f\left(\x, \y^{*}(\x)\right)
\quad \text {s.t.} \quad \y^{*}(\x) = \arg \min _{\y \in \mathbb{R}^{d_l}} g(\x, \y), 
\end{gather}
where $f: \mathbb{R}^{d_u} \times \mathbb{R}^{d_l} \to \mathbb{R}$ and $g: \mathbb{R}^{d_u} \times \mathbb{R}^{d_l} \to \mathbb{R}$ denote the UL and LL objectives, respectively.
This framework has been widely adopted across a broad range of machine learning applications, including hyperparameter optimization~\cite{bao2021stability, franceschi2018bilevel}, meta-learning~\cite{rajeswaran2019meta, ji2020convergence}, reinforcement learning~\cite{hong2023two, shen2025principled}, neural architecture search~\cite{lian2019towards, hu2020tf}, adversarial training~\cite{tian2021alphagan, zhang2022revisiting, li2019learning}, data hyper-cleaning~\cite{franceschi2018bilevel, shaban2019truncated}, and dictionary learning~\cite{lecouat2020flexible, mairal2011task}.

Most existing methods for solving Problem~(\ref{eq: BLO}) are developed under the assumption that the LL objective is explicitly available and can be differentiated, queried, or optimized by the UL learner.  
However, this assumption may fail in many settings where the LL solution is produced by an external procedure whose objective or internal mechanism is inaccessible.  
For example, in adversarial training, the UL problem updates the model parameters, while the LL problem generates adversarial examples or perturbations against the current model.  
When this LL procedure is implemented through a black-box attack, simulator, or learned generator, the UL training algorithm may only observe the resulting adversarial examples, without direct access to the underlying LL objective, a differentiable description of the LL objective (i.e., gradient, Hessian, or Jacobian), or the internal optimization procedure that produce them.

Motivated by applications where the LL response is observable but the objective or mechanism generating it is unavailable to the UL learner, we study a new problem setting, which we call \emph{LL agnostic} bilevel optimization.  
In this setting, the LL objective $g$ still defines the target response $\y^*(\x)$ in Eq.~(\ref{eq: BLO}), but the UL algorithm does not have access to $g$, including its function values ({\em zeroth-order} information), gradients ({\em first-order} information), and Hessian/Jacobian information ({\em second-order} information).  
Instead, for a UL variable $\x$, the algorithm only observes an approximate LL solution $\bar{\y}(\x)$, which may be produced by a black-box solver, simulator, or learned predictor.

This LL-agnostic setting goes beyond the scope of existing bilevel optimization algorithms.
When the LL problem is strongly convex, classical bilevel optimization methods update $\x$ using the gradient of the UL objective (UL-gradient), which can be computed using the chain rule as:
\begin{align}\label{eq:exact_hypergradient}
  & \nabla \ell(\x) =  \nabla_\x f(\x, \y^\ast(\x)) + [\nabla_{\x} \y^\ast(\x)]^\top \nabla_\y f(\x, \y^\ast(\x)).
\end{align}
Approximate implicit differentiation (AID) methods~\cite{hong2023two, ghadimi2018approximation, ji2021bilevel, chen2021tighter, arbel2021amortized, dagreou2022framework, khanduri2021near,pedregosa2016hyperparameter} estimate the Jacobian $\nabla_{\x} \y^\ast(\x)$ in Eq.~(\ref{eq:exact_hypergradient}) by applying the implicit function theorem~\cite{rudin1976principles}.
This requires derivative information from the LL objective, i.e., $\nabla^2_{\x\y} g$ and $\nabla^2_{\y\y} g$, to construct the resulting hypergradient.
Iterative differentiation (ITD) methods~\cite{franceschi2018bilevel, shaban2019truncated, domke2012generic, maclaurin2015gradient, franceschi2017forward, grazzi2020iteration, mackay2019self} instead differentiate through the LL optimization trajectory, and therefore require access to the LL update rule or computational graph.
Both classes of methods are {\em incompatible} with the LL-agnostic setting, where the UL algorithm {\em cannot} evaluate or differentiate the LL objective and {\em cannot} necessarily backpropagate through the procedure that returns $\bar{\y}(\x)$.

Existing Hessian/Jacobian-free and derivative-free bilevel methods are also not directly applicable.  
Although these methods avoid second-order derivative computations, they assume an information model in which the LL objective is accessible in some form.  
Hessian/Jacobian-free first-order methods~\cite{yang2023achieving, kwon2023fully, kwon2024on, liu2022bome, shen2023penalty} avoid explicit second-order derivatives, but still rely on gradients of the LL objective or reformulations based on LL optimality conditions. 
Zeroth-order bilevel methods~\cite{song2020esmaml, aghasi2025fully, aghasi2025optimal, rando2026zoba} typically construct finite-difference or smoothing-based estimators using function-value queries of the UL and LL objectives. 
A different line of derivative-free work considers direct-search methods that assume access to a black-box LL oracle returning approximate LL solutions and optimize the resulting inexact bilevel objective directly through function-value comparisons~\cite{diouane2024inexact}.
Such methods avoid differentiating the LL problem, but treat the induced UL objective as a derivative-free black box rather than exploiting its available first-order structure.
In contrast, the LL-agnostic setting provides no function-value or gradient access to $g$, while the UL objective $f$ remains differentiable and accessible. Our goal is therefore not to treat the entire bilevel problem as a black box.
Instead, we exploit the available gradients of $f$ and use only the observable LL response $\bar{\y}(\x)$ to approximate the missing sensitivity $\nabla_{\x}\bar{\y}(\x)$.
Thus, the key difficulty is not merely the absence of Hessian/Jacobian information or derivative information from the induced bilevel objective, but the absence of a usable LL objective altogether.

To address this difficulty, we propose \algcd~(\ul{L}ower-l\ul{e}vel A\ul{g}n\ul{o}stic \ul{Bi}level \ul{O}ptimization), a {\em coordinate-wise pseudo-hypergradient} method that uses the observable LL response map $\x \mapsto \bar{\y}(\x)$ to estimate the missing Jacobian $\nabla_{\x} \y^\ast(\x)$ in Eq.~(\ref{eq:exact_hypergradient}). 
Once $\bar{\y}(\x)$ is used as a tractable proxy for $\y^*(\x)$, the UL-gradient in Eq.~(\ref{eq:exact_hypergradient}) can be approximated by replacing $\y^*(\x)$ with $\bar{\y}(\x)$. 
The remaining challenge is to estimate the Jacobian $\nabla_{\x}\bar{\y}(\x)$ without differentiating through $g$ or through the mechanism that produces $\bar{\y}$. 
\algcd~addresses this challenge by updating one UL coordinate at a time.
Specifically, for each UL coordinate, \algcd~constructs a decision vector subject to 1-sparse updates, i.e., restricting modifications to a single coordinate per step, thereby isolating the corresponding Jacobian column needed in the UL gradient computation.
This coordinate-wise Jacobian estimate is then combined with the UL gradients (i.e., $\nabla_{\x}f$ and $\nabla_{\y}f$) to form a pseudo-hypergradient.
As a result, \algcd~requires only gradients of the UL objective and queries to the approximate LL solution map without needing $g$, $\nabla g$, $\nabla^2_{\x\y}g$, or $\nabla^2_{\y\y}g$, making it applicable to LL-agnostic setting.

The coordinate-wise design is important for both statistical and algorithmic reasons. 
A finite difference based on a full-parameter change in $\x$ only provides one directional derivative of $\bar{\y}$ along the chosen displacement, and therefore cannot identify the entire Jacobian matrix information needed to form a full UL-gradient without additional perturbations or structural assumptions. 
In contrast, \algcd~isolates one Jacobian column at a time, which aligns naturally with coordinate-wise UL updates and avoids mixing the effects of multiple coordinates in a single response difference. 
Moreover, \algcd~uses finite differences only to estimate the missing Jacobian within the UL-gradient, while retaining the available first-order gradients of the UL objective. 
This distinguishes it from the zeroth-order approaches that approximate the entire UL-gradient through function-value differences of the composite UL objective $f(\x,\bar{\y}(\x))$, which are typically more sensitive to noise and dimensionality~\cite{duchi2015optimal}.

Our main contributions are summarized as follows.
\begin{itemize}[leftmargin=3 mm]
\vspace{-.1in}

\item We formulate and analyze a LL agnostic bilevel optimization setting in which the LL objective is unavailable to the UL algorithm, and the only available LL-side information is an approximate solution $\bar{\y}(\x)$. 
This information model distinguishes the problem from both classical gradient-based bilevel optimization and standard zeroth-order bilevel optimization.

\item We develop \algcd, a coordinate-wise pseudo-hypergradient algorithm that estimates the Jacobian of the approximate LL solution (i.e., $\nabla _{\x}\bar{\y}$ in Eq.~(\ref{eq:exact_hypergradient})) {\em using only differences between LL responses}. 
This method avoids differentiating through the LL objective or the procedure that generates $\bar{\y}(\x)$, and is therefore applicable to the LL-agnostic setting.

\item We establish convergence guarantees for \algcd~with nonconvex UL objectives in both deterministic and stochastic settings.
The analysis does not require strong convexity of the LL objective (LLSC); instead, it relies on an LL error-bound/Polyak–Łojasiewicz-type condition.
\algcd~achieves an $\mathcal{O}(T^{-1})$ convergence rate in the deterministic setting and an $\mathcal{O}(T^{-1/2})$ rate in the stochastic setting when the LL approximation error decreases geometrically.
Despite operating in the more challenging LL-agnostic setting, these rates match the state-of-the-art deterministic and stochastic convergence rates achieved by standard bilevel methods under the LLSC assumption.

\item We evaluate the numerical performance of \algcd~on three tasks: deep hyper-representation, adversarial training, and data weighting for LLM alignment. The results show that, despite having no access to the LL objective, \algcd~achieves competitive or superior performance compared with baselines that use LL objective information during training.

\end{itemize}


\section{Related Work} \label{sec: RelatedWork}

In this section, we provide an overview on three closely related lines of research to place our work in comparative perspectives.

{\bf 1) Gradient-based Bilevel Optimization:}
Gradient-based bilevel optimization methods typically compute or approximate the UL-gradient of the value function $\ell(\x)=f(\x,\y^*(\x))$.
Two major classes of methods are approximate implicit differentiation (AID) and iterative differentiation (ITD).  
AID-based methods use the implicit function theorem to approximate the implicit gradient of $\ell(\x)$~\cite{hong2023two, ghadimi2018approximation, ji2021bilevel, chen2021tighter, arbel2021amortized, dagreou2022framework, khanduri2021near,pedregosa2016hyperparameter}. 
These methods generally require LL derivative information, such as $\nabla^2_{\x\y}g$ and $\nabla^2_{\y\y}g$, or Hessian-vector and Jacobian-vector products derived from the LL objective.
ITD-based methods instead approximate the UL-gradient by differentiating through the unrolled LL optimization trajectory~\cite{domke2012generic, maclaurin2015gradient, franceschi2017forward, franceschi2018bilevel, shaban2019truncated, grazzi2020iteration, mackay2019self}.
As a result, ITD methods require access to the LL update rule or computational graph, since the UL-gradient is obtained by backpropagating through the LL solver.

These approaches are effective when the LL objective is available and differentiable, but they are inapplicable for the LL-agnostic setting studied in this paper.
In our setting, the UL learner cannot query the LL objective $g$, compute $\nabla^2_{\x\y}g$ and $\nabla^2_{\y\y}g$, form Hessian/Jacobian-vector products, or backpropagate through the procedure that produces the approximate LL solution $\bar{\y}(\x)$. 
Therefore, neither AID nor ITD can be directly applied. 
In contrast, \algcd~requires only gradients of the UL objective and queries to the observable LL response map $\x\mapsto \bar{\y}(\x)$. 
This distinction is also reflected in the convergence analysis: many gradient-based bilevel methods rely on LL strong convexity to control the solution map and hypergradient approximation errors, whereas our analysis uses an LL error-bound condition together with an explicit approximation-error sequence for $\bar{\y}(\x)$.

{\bf 2) Hessian/Jacobian-free First-order Bilevel Optimization:}
A recent line of work develops Hessian/Jacobian-free first-order bilevel methods to reduce the computational burden of second-order hypergradient estimation. 
These methods avoid explicitly computing LL Hessian or Jacobian matrices and instead replace second-order quantities with first-order approximations, optimality-based constraints, or penalty/value-function reformulations. 
For example, \cite{yang2023achieving} approximates second-order information using first-order gradient information. 
Other methods reformulate the bilevel problem through LL optimality conditions and penalty/value-function-based objectives so that the resulting algorithms only require first-order gradient information~\cite{kwon2023fully, kwon2024on, liu2022bome, shen2023penalty}.

Although these methods avoid explicit second-order derivatives, they still assume access to the LL objective in a usable form. 
In particular, they typically require LL gradients, LL optimality residuals, or penalty/value functions involving the LL objective.
Hence, they remain incompatible with the LL-agnostic setting.
The challenge in our setting is not merely that Hessian/Jacobian information is unavailable; rather, the LL objective itself is unavailable to the UL learner. 
\algcd~therefore differs from Hessian/Jacobian-free first-order bilevel algorithms by estimating the missing Jacobian $\nabla _{\x}\bar{\y}$ directly from differences of approximate LL solutions, without using $g$ or its gradients.

{\bf 3) Derivative-free/Zeroth-order Bilevel Optimization:}
Derivative-free/zeroth-order bilevel optimization studies settings where derivative information is unavailable and algorithms rely on function-value queries or direct evaluations of the induced bilevel objective. 
Several works develop finite-difference, evolution-strategy, or Gaussian-smoothing estimators for bilevel optimization.
For example, evolution-strategy-based methods have been used to avoid second-order derivatives in meta-learning~\cite{song2020esmaml}. 
Gaussian-smoothing-based methods estimate derivatives of both UL and LL objectives from noisy function values~\cite{aghasi2025fully, aghasi2025optimal}. 
More recent single-loop zeroth-order bilevel methods further improve the efficiency of finite-difference UL-gradient approximation~\cite{rando2026zoba}.
Beyond gradient-estimation-based approaches, derivative-free direct-search methods have also been studied for bilevel optimization.
In particular, \cite{diouane2024inexact} develop inexact direct-search schemes that use a fixed-accuracy black-box LL solver and optimize the resulting approximate bilevel objective through function-value comparisons, without explicitly constructing gradient estimators.

Our method is related to this literature because it also uses finite differences, but both the information model and the role of function evaluations are different.
Standard zeroth-order bilevel methods assume access to function values of the objectives, including the LL objective $g$, and use finite differences or smoothing to estimate gradients of the UL and LL objectives. 
Likewise, direct-search bilevel methods treat the induced UL objective as a derivative-free black box and make optimization decisions based on its function values.
In contrast, the LL-agnostic setting provides {\em no function-value access to $g$.} 
At the same time, the UL objective $f$ remains differentiable and accessible. 
Therefore, \algcd~does not treat either level, or the induced bilevel objective, as a fully black-box optimization problem.
Instead, it preserves the chain-rule structure of the UL-gradient by using $\nabla_{\x}f$ and $\nabla_{\y}f$ directly, while applying finite differences only to estimate the missing Jacobian $\nabla_{\x}\bar{\y}(\x)$ of the approximate LL solution. 
This targeted use of finite differences distinguishes \algcd~from both gradient-estimation-based zeroth-order methods and direct-search-based derivative-free methods, and is central to its LL agnostic design. 
Since function-value-based zeroth-order estimators often incur additional sampling and dimension-dependent errors~\cite{duchi2015optimal}, retaining the available UL gradients in \algcd~can lead to a more accurate and less noisy UL-gradient approximation.


\section{The \algcd~Algorithm} \label{sec: Algorithm_CD}


In this section, we propose \algcd, a coordinate-wise pseudo-hypergradient method for LL agnostic bilevel optimization. 
The main idea is to estimate the Jacobian of the observable LL response map $\x \mapsto \bar{\y}(\x)$ by updating one UL coordinate at a time and measuring the resulting change in the approximate LL solution.
This construction enables \algcd~to approximate the missing Jacobian information of $\bar{\y}$ without accessing the LL objective $g$, its gradient $\nabla g$, or its second-order derivatives, i.e., $\nabla_{\x\y}^2 g$ and $\nabla_{\y\y}^2 g$.

The \algcd~algorithm is summarized in Algorithm~\ref{alg_cd}.
For each UL coordinate, \algcd~constructs a decision vector subject to 1-sparse updates, which restricts modifications to a single coordinate per step, thereby isolating the corresponding Jacobian column needed in the UL gradient computation.
Specifically, at iteration $t$ and for each coordinate $j\in\{1,\ldots,d_u\}$, define the coordinate displacement as
\begin{align}\label{eq: cd-point}
    s_t^j \triangleq x_{t-1}^j-x_t^j,\quad
    \widetilde{\x}_t^j \triangleq \x_{t-1} - s_t^j \e_j,
\end{align}
\begin{wrapfigure}{r}{0.52\textwidth}
\vspace{-0.2in}
\begin{minipage}[t]{1\linewidth}
\begin{algorithm}[H]
\caption{The \algcd~algorithmic framework.}\label{alg_cd}
\begin{small}
\begin{algorithmic}
\State \textbf{Input:} Initial points $\x_{-1}, \x_0 \in \mathbb{R}^{d_u}$, and upper-level step sizes $\{\alpha_t\}_{t=0}^{T-1}$
\For{$t=0,\dots, T-1$}
    \State Receive the lower-level approximation $\bar{\y}(\x_{t-1})$
    \For{$j=1,2,\dots,d_u$}
        \State Form the point $\widetilde{\x}_t^j$ as in Eq.~(\ref{eq: cd-point})
        \State Receive the lower-level approximation $\bar{\y}(\widetilde{\x}_t^j)$
        \State Compute pseudo-hypergradient $h_{t}^{f,j}$ via Eq.~(\ref{eq:h_tf_cd}) in deterministic setting or Eq.~(\ref{eq:h_tf_cd_sto}) in stochastic setting
        \State Update $x_{t+1}^j = x_{t}^j - \alpha_t \big[h_{t}^{f,j}\big]_j$
    \EndFor
\EndFor
\end{algorithmic}
\end{small}
\end{algorithm}
\end{minipage}
\vspace{-.12in}
\end{wrapfigure}
where $\e_j$ denotes the $j$-th canonical basis vector in $\mathbb{R}^{d_u}$.
The point $\widetilde{\x}_t^j$ differs from $\x_{t-1}$ only in the $j$-th coordinate. 
Therefore, the difference $\bar{\y}(\x_{t-1})-\bar{\y}(\widetilde{\x}_t^j)$ captures how the approximate LL response changes when only the $j$-th UL coordinate is updated.
This provides a one-dimensional finite-difference estimate of the $j$-th column of the Jacobian $\nabla_{\x}\bar{\y}(\x)$:
\begin{align*}
    \frac{\bar{\y}(\x_{t-1})-\bar{\y}(\widetilde{\x}_t^j)}{s_t^j}
    \approx \nabla_{\x}\bar{\y}(\x_{t})\e_j .
\end{align*}
Throughout the paper, we assume that $s_t^j\neq 0$ to ensure that the finite-difference estimate is well-defined.
This column-wise estimate is exactly the quantity needed for the $j$-th UL-gradient component, because
\begin{align*}
    \left[\nabla_{\x}\bar{\y}(\x)^\top \nabla_{\y}f(\x,\bar{\y}(\x))\right]_j
    = \left\langle \nabla_{\x}\bar{\y}(\x)\e_j,\nabla_{\y}f(\x,\bar{\y}(\x)) \right\rangle .
\end{align*}

This explains why \algcd~uses coordinate-wise response differences rather than a single full-parameter response difference.
A full-vector perturbation would give
\begin{align*}
    \bar{\y}(\x_{t-1})-\bar{\y}(\x_t)
    \approx \nabla_{\x}\bar{\y}(\x_{t})(\x_{t-1}-\x_t),
\end{align*}
which is only one directional derivative of the response map.
After contraction with $\nabla_{\y}f$, it yields one scalar projection
\begin{align*}
    \left\langle \bar{\y}(\x_{t-1})-\bar{\y}(\x_t), \nabla_{\y}f(\x_t, \bar{\y}(\x_t)) \right\rangle
    \approx \left\langle \x_{t-1}-\x_t, \nabla_{\x}\bar{\y}(\x_{t})^\top \nabla_{\y}f(\x_t, \bar{\y}(\x_t)) \right\rangle,
\end{align*}
not the full vector $\nabla_{\x}\bar{\y}(\x_{t})^\top \nabla_{\y}f(\x_t, \bar{\y}(\x_t))$ needed to update all UL coordinates.
Recovering the full vector from full-parameter perturbations would require multiple independent perturbations or a random-direction estimator, which introduces additional sampling error and dimension-dependent variance.
The coordinate-wise construction instead isolates each component directly, leading to a more accurate UL-gradient estimate.

\paragraph{Deterministic pseudo-hypergradient.}
In the deterministic setting, the coordinate-wise pseudo-hypergradient is defined as
\begin{align} \label{eq:h_tf_cd}
    h_{t}^{f,j}=\nabla_{\x} f  \left(  \widetilde{\x}_t^j, \bar{\y}(\widetilde{\x}_t^j) \right) +\bigg\langle \frac{\bar{\y}(\x_{t-1})-\bar{\y}(\widetilde{\x}_t^j)}{ s_t^j},\nabla_{\y} f(\widetilde{\x}_t^j, \bar{\y}(\widetilde{\x}_t^j)) \bigg\rangle \e_{j}.
\end{align}
Only the $j$-th component $[h_t^{f,j}]_j$ is used to update the coordinate $x_t^j$.
The first term in Eq.~(\ref{eq:h_tf_cd}) is the direct UL gradient with respect to $\x$.
The second term estimates the indirect effect of changing $\x$ through the LL response map.
The Jacobian $\nabla_{\x}\bar{\y}(\x)$ is estimated through a one-dimensional finite difference of $\bar{\y}(\cdot)$.
This coordinate-wise construction is the key mechanism that allows \algcd~to construct a pseudo hypergradient using only LL responses rather than derivatives of the unknown LL objective. It is also important to distinguish Eq.~(\ref{eq:h_tf_cd}) from a finite-difference approximation of the UL-gradient using composite UL objective $f(\x,\bar{\y}(\x))$.
One could estimate the $j$-th derivative of the entire UL-gradient by
\begin{align*}
    \frac{ f(\x_{t-1},\bar{\y}(\x_{t-1}))-f(\widetilde{\x}_t^j,\bar{\y}(\widetilde{\x}_t^j))
    }{s_t^j},
\end{align*}
but this would treat the entire UL objective as a black box.
Such an estimator would discard the available UL gradients $\nabla_{\x}f$ and $\nabla_{\y}f$, mix the direct UL variation and the LL-response variation into a single scalar difference, and can amplify errors through the subtraction of inaccurate function values.
Our LL-agnostic setting is different: the UL objective $f$ remains differentiable and accessible, while only the LL objective $g$ and its derivatives are unavailable.
Therefore, \algcd~preserves the chain-rule structure of the UL-gradient: it differentiates $f$ directly and applies finite differences only to estimate the missing Jacobian $\nabla_{\x}\bar{\y}$. 
This targeted use of finite differences leads to a more accurate and less noisy UL-gradient surrogate.

\paragraph{Stochastic pseudo-hypergradient.}
We further consider the stochastic setting, where the UL and LL objectives are given by expectations:
\begin{gather*}
f\left(\x, \y^{*}(\x)\right) \triangleq \mathbb{E}_{\xi \sim \pi_{f}} \left[ f\left(\x, \y^{*}(\x) ; \xi\right)\right],
\quad g(\x, \y) \triangleq \mathbb{E}_{\zeta \sim \pi_{g}}[g(\x, \y ; \zeta)],
\end{gather*}
where $\xi \sim \pi_f$ and $\zeta \sim \pi_g$ denote samples from the UL and LL data distributions, respectively.
When only stochastic estimates are available, \algcd~replaces the exact quantities in Eq.~(\ref{eq:h_tf_cd}) with mini-batch estimates:
\begin{align} \label{eq:h_tf_cd_sto}
    \widehat{h}_{t}^{f,j}=\nabla_{\x} f \left( \widetilde{\x}_t^j, \bar{\y}(\widetilde{\x}_t^j);\mathcal{B}_t^{f_x} \right) \!+\!\bigg\langle \frac{\bar{\y}(\x_{t-1};\mathcal{B}_{t-1}^{y})-\bar{\y}(\widetilde{\x}_t^j;\mathcal{B}_t^{y,j})}{ s_t^j},\nabla_{\y} f(\widetilde{\x}_t^j, \bar{\y}(\widetilde{\x}_t^j);\mathcal{B}_t^{f_y,j}) \bigg\rangle \e_{j},
\end{align}
where $\mathcal{B}_t^{f_x}$ and $\mathcal{B}_t^{f_y,j}$ denote mini-batches used to estimate the UL gradients with respect to $\x$ and $\y$, respectively. 
$\mathcal{B}_t^{y,j}$ and $\mathcal{B}_{t-1}^{y}$ denote mini-batches used to obtain the stochastic LL responses at $\widetilde{\x}_t^j$ and $\x_{t-1}$.
They are introduced only to characterize the stochastic approximation error and are not required to be known by the UL learner.
This estimator highlights the three main sources of error in the stochastic LL-agnostic setting: the gradient approximation error from the UL sampling, the approximation error from replacing the optimal LL solution $\y^*(\x)$ with $\bar{\y}(\x)$, and the Jacobian-estimation error from approximating $\nabla_{\x}\bar{\y}(\x)$ through finite differences of stochastic LL responses.


\section{Theoretical Performance Analysis} \label{sec: ConvergenceAnalysis}

In this section, we establish the convergence guarantees for \algcd~in both deterministic and stochastic settings. 
A key feature of our analysis is that it does not require the LL objective to be strongly convex. 
Instead, we rely on an error-bound/PL-type condition for the LL problem.

\subsection{Assumptions}

We first state a set of assumptions that are needed to establish the convergence of \algcd.
Although \algcd~does not require direct access to the LL objective $g$, the assumptions below characterize the underlying bilevel problem and the accuracy required of the LL solution approximation.

\begin{assumption}[UL Objective]\label{asmp_up}
    Let $\nabla f(\x, \y)=  \nabla_\x f(\x, \y) + [\nabla_{\x} \y(\x)]^\top \nabla_\y f(\x, \y)$.
    The UL objective $f(\x,\y)$ satisfies the following conditions.\\
    \textit{1)} The gradient map $(\x, \y) \mapsto \nabla f(\x, \y)$ is $L_{f}$-Lipschitz continuous for some $L_{f}\ge 0$, i.e., $\left\|\nabla f(\x_1,\y_1)-\nabla f(\x_2,\y_2)\right\| \leq L_{f}(\left\|\x_1-\x_2\right\|+\left\|\y_1-\y_2\right\|)$.\\
    \textit{2)} For all $\left( \x,\y \right)\in \mathbb{R}^{d_u}\times \mathbb{R}^{d_l}$, we have $\left\| \nabla _{\y}f\left( \x,\y \right) \right\|\le B_{f_{y}}$ for some $B_{f_{y}}\ge 0$.  
\end{assumption}

Assumption~\ref{asmp_up} is standard in nonconvex bilevel optimization~\cite{hong2023two, ghadimi2018approximation, khanduri2021near}. 
The Lipschitz condition controls the smoothness of the UL objective, while the boundedness of $\nabla_{\y} f$ ensures that errors in the LL response do not propagate arbitrarily into the UL hypergradient.

\begin{assumption}[LL Objective]\label{asmp_low}
    The LL objective $g(\x,\y)$ satisfies the following conditions.\\
    \textit{1)} For all $\x \in \mathbb{R}^{d_u}$, the LL problem admits a unique solution $\y^*\left( \x \right)$, which is differentiable.\\
    \textit{2)} For all $\left( \x,\y \right)\in \mathbb{R}^{d_u}\times \mathbb{R}^{d_l}$, we have $\left\|\y-\y^*\left( \x \right)\right\|\le \kappa\left\| \nabla _\y g\left( \x,\y \right) \right\|$ for some $\kappa\ge 0$.\\
    \textit{3)} For all $\y \in \mathbb{R}^{d_l}$, the map $\x \mapsto \nabla_{\y} g(\x, \y)$ is $L_{g_y}$-Lipschitz continuous for some $L_{g_y}\ge 0$, i.e., $\left\| \nabla _\y g\left( \x_1,\y\right)-\nabla _\y g\left( \x_2,\y\right) \right\|\le L_{g_y}\left\| \x_1-\x_2 \right\|$, $\mathbf{x}_1,\mathbf{x}_2\in \mathbb{R}^{d_u}$.
\end{assumption}

Rather than assuming that the LL objective is $\mu_g$-strongly convex, as is commonly adopted in the bilevel optimization literature~\cite{ghadimi2018approximation, hong2023two,khanduri2021near, liu2022interact, qiu2023diamond}, Assumption~\ref{asmp_low} imposes an LL error-bound/PL-type condition together with uniqueness of the LL solution.
The PL-type condition is weaker than strong convexity and can hold for certain nonconvex objectives. 
Importantly, the uniqueness requirement does not imply strict or strong convexity; nonconvex LL objectives may still admit a unique, isolated solution while satisfying the stated PL-type error bound. Therefore, our analysis is not restricted to convex or strongly convex LL problems. 
As a special case, when $g(\x,\cdot)$ is $\mu_g$-strongly convex, the LL PL-type condition is satisfied with $\kappa=1/\mu_g$. 
Assumptions~\ref{asmp_low} and~\ref{asmp_up} are used to establish the Lipschitz continuity of $\nabla \ell(\x)$.

Assumption~\ref{asmp_low}(2) gives $\left\|\bar{\mathbf{y}}(\mathbf{x})-\mathbf{y}^*(\mathbf{x})\right\| \leq \kappa \epsilon_t$, so $\delta_t=\kappa \epsilon_t$. 
A fixed LL budget yields an approximation-dependent neighborhood, whereas an increasing budget can make $\delta_t$ decrease and enable asymptotic stationarity.
Note that the UL learner does {\bf not} need to access $g$ or know $\delta_t$.

\begin{assumption}[LL Approximation]\label{asmp_y}
    The approximate LL response map $\bar{\y}(\x)$ satisfies the following conditions.\\
    \textit{1)} For all $t\!\in\! \left\{ 0,1,\ldots,T\!-\!1  \right\}$ and every $\x_t \!\in\! \mathbb{R}^{d_u}$, we have $\left\| \bar{\y}\left(\x_t\right)\!-\!\y^*\left(\x_t\right) \right\|\!\le\! \delta_t$ for some $\delta_t\!\ge\! 0$. \\
    \textit{2)} The Jacobian map $\x\mapsto \nabla_{\x}\bar{\y}(\x)$ is $L_{y}$-Lipschitz continuous for some $L_{y}\ge 0$, and 
    $\left\| \nabla _{\x}\bar{\y}\left( \x \right)\right\|\le B_y$ for some $B_y\ge 0$.
\end{assumption}


Assumption~\ref{asmp_y} captures \algcd’s two error sources: LL approximation, and finite-difference Jacobian estimation. 
Their global and uniform forms are sufficient conditions for analysis, while in practice they need only hold locally along the iterates and queried response segments.
Assumption~\ref{asmp_y}(1) follows naturally from practical LL stopping criteria. 
If an external LL solver returns $\bar{\mathbf{y}}(\mathbf{x})$ satisfying $\left\|\nabla_\mathbf{y} g(\mathbf{x}, \bar{\mathbf{y}}(\mathbf{x}))\right\| \leq \epsilon_t$, 
Assumption~\ref{asmp_y}(2) is reasonable when the LL oracle induces a smooth and stable input-output map. 
For example, it holds for an exact isolated, nondegenerate LL solution when the LL Hessian is uniformly nonsingular and the relevant second derivatives of $g$ are bounded and Lipschitz. 
It also holds when $\bar{\mathbf{y}}(\mathbf{x})$ is generated by a fixed number of smooth solver updates whose derivatives are bounded and Lipschitz on the region visited by the iterates.
Note that the UL learner {\bf need not} evaluate the Jacobian or know $B_y$ and $L_y$; they only characterize the black-box map’s regularity.

For the stochastic setting, we further impose the following assumptions on the stochastic estimators.

\begin{assumption}[Stochastic Gradients]\label{asmp_sg}
    For any $\left( \x,\y \right) 
    \in \mathbb{R}^{d_u} \times \mathbb{R}^{d_l}$ and data batches $\mathcal{B}^{f_{x}}$, $\mathcal{B}^{f_{y}}$, and $\mathcal{B}^{y}$, 
    the stochastic gradients of the UL objective are unbiased and have bounded variances:  
    \begin{gather*}
        \mathbb{E}[\|\nabla_{\x} f(\x, \y ; \mathcal{B}^{f_{x}})-\nabla_{\x}  f(\x, \y)\|^{2}] \leq \sigma_{f_{x}}^{2}, \quad
        \mathbb{E}[\|\nabla_{\y} f(\x, \y ; \mathcal{B}^{f_{y}})-\nabla_{\y}  f(\x, \y)\|^{2}] \leq \sigma_{f_{y}}^{2}.
    \end{gather*}
    In addition, the stochastic approximation used for estimating the Jacobian satisfies, for every $j$-th coordinate,
    \begin{gather*}
        \mathbb{E}\left[ \left\| \left( \nabla _{\x}\bar{\y}(\x;\mathcal{B}^{y})-\nabla _{\x}\bar{\y}(\x) \right)\e_j \right\|^2 \right] \leq \sigma_{y}^{2}.
    \end{gather*}
\end{assumption}

In Assumption~\ref{asmp_sg}, the variance terms vanish in the full-batch setting, and the Jacobian condition is an analytical mean-square regularity condition rather than an additional computational requirement. 

Finally, we use the following stationarity criterion.

\begin{definition}[$\epsilon$-Stationarity] \label{Def: StationaryPt} 
A point $\x$ is an $\epsilon$-stationary solution if $\mathbb{E}\big[ \left\| \nabla \ell\left( \x \right) \right\|^{2} \big]\le \epsilon$, where the expectation is taken over all randomness of the algorithm.
\end{definition}

\subsection{Main Convergence Results}

We now state the main convergence result of \algcd~for non-convex $\ell\left( \x \right)$ in the deterministic setting, as established in Theorem~\ref{thm_cd}:

\begin{theorem}[Deterministic Setting]\label{thm_cd}
    Under Assumptions~\ref{asmp_up}--\ref{asmp_y}, choose constant step-sizes $\alpha_t=\alpha$ for all $t \in \{0,1, \ldots, T-1\}$, with $0< \alpha\leq\bar{\alpha}$, where $\bar{\alpha}$ depends on $L_l$, $d_u$, $d_l$, $B_{f_y}$ and $L_y$, and is specified in~\cref{apx:thm_cd}.
    Then, the iterates generated by \algcd~satisfy:
    \begin{align*}
        &\frac{1}{T}  \sum_{t=0}^{T-1} \left\|\nabla \ell\left(\x_t\right)\right\|^2 = \mathcal{O}\left( \frac{ \Delta_0 }{T\alpha} \right)+\mathcal{O}\bigg( d_u\frac{1}{T}  \sum_{t=0}^{T-1}\delta_{t+1}^2 \bigg),
    \end{align*}
   where $\Delta_0  =  \left( \ell\left( \x_0 \right)-\ell^* \right)+\|\nabla \ell\left( \x_{0} \right)-\sum_{j=1}^{d_u}[h_0^{f, j}]_j \e_j\|^2$.
\end{theorem}

The proof of \cref{thm_cd} is provided in~\cref{apdx: Proof_Deterministic_CD}. 
Theorem~\ref{thm_cd} shows that the convergence error consists of two terms: the standard optimization error $\mathcal{O}(\Delta_0/(T\alpha))$ and an additional bias term induced by the LL approximation error. 
This bias is controlled by the average squared error of the approximate LL solutions.
Hence, the asymptotic convergence behavior depends on how the LL approximation error evolves over the course of optimization.
If the LL approximation error $\delta_t$ is a non-vanishing constant, i.e., $\delta_t=\delta>0$, for a fixed admissible step size, the optimization error vanishes as $T \to \infty$, but a residual bias remains. 
This does not imply divergence; rather, \algcd~is guaranteed to reach an $\mathcal{O}\left( d_u\delta^2 \right)$-stationary neighborhood. Such an error floor is expected when the LL oracle has persistent approximation bias, since exact stationarity of the original objective cannot generally be guaranteed from permanently inaccurate LL responses.

Moreover, if the accuracy of the returned LL responses improves sufficiently fast, the corresponding bias diminishes with $T$. 
Such behavior can naturally arise when the external LL oracle is progressively refined, for example, by increasing its computational budget or tightening its stopping tolerance across iterations. 
Importantly, \algcd~does not require the UL learner to know or control the refinement mechanism; this condition only characterizes the quality of the responses returned by the external LL oracle. 
The following corollary makes this dependence explicit and establishes the resulting convergence rate under a decreasing LL approximation error.

\begin{corollary}\label{cor: diminish_delta}
    Under the same conditions as in Theorem~\ref{thm_cd}, choose $\alpha= \Theta\left( 1 \right)$ and assume that the LL approximation error decreases geometrically, i.e., $\delta_{t+1}^2\le \left( 1-\gamma \right)\delta_{t}^2$ for some $0<\gamma<1$.
    Then, the iterates generated by \algcd~satisfy: $\frac{1}{T}  \sum_{t=0}^{T-1} \left\|\nabla \ell\left(\x_t\right)\right\|^2 = \mathcal{O}\left( T^{-1} \right)$.
\end{corollary}

Corollary~\ref{cor: diminish_delta} shows that, when the LL approximation error decreases over time, \algcd~achieves an $\mathcal{O}\left( T^{-1} \right)$ convergence rate for reaching a stationary solution. 
This rate matches the state-of-the-art deterministic convergence rate achieved by standard bilevel methods under stronger LL strong-convexity assumptions~\cite{ghadimi2018approximation, ji2021bilevel}.

We next present the main convergence result of \algcd~for non-convex $\ell\left( \x \right)$ in the stochastic setting, as given in Theorem~\ref{thm_cd_sto}:

\begin{theorem}[Stochastic Setting]\label{thm_cd_sto}
    Under Assumptions~\ref{asmp_up}--\ref{asmp_sg}, choose constant step-sizes $\alpha_t=\alpha$ for all $t \in \{0,1, \ldots, T-1\}$, with $0<\alpha\leq\bar{\alpha}$, where $\bar{\alpha}$ depends on $L_l$, $d_u$, $d_l$, $B_{f_y}$ and $L_y$, and is specified in~\cref{apx:thm_cd_sto}.  
    Then, the iterates generated by \algcd~satisfy:
    \begin{align*}
        &\frac{1}{T} \sum_{t=0}^{T-1} \mathbb{E}\left[ \left\|\nabla \ell\left(\x_t\right)\right\|^2 \right] = \mathcal{O}\left( \frac{ \Delta_0 }{T\alpha} \right)+\mathcal{O}\bigg( d_u \frac{1}{T}  \sum_{t=0}^{T-1}\delta_{t+1}^2+\alpha d_u\big(\sigma_{f_{x}}^{2}+\sigma_y^2 + \sigma_{f_{y}}^{2}+\sigma_y^2\sigma_{f_{y}}^{2} \big) \bigg),
    \end{align*}
   where $\Delta_0  =  \left( \ell\left( \x_0 \right)-\ell^* \right)+\|\nabla \ell\left( \x_0 \right)-\sum_{j=1}^{d_u}[h_0^{f, j}]_j \e_j\|^2$.
\end{theorem}

The proof of \cref{thm_cd_sto} is provided in~\cref{apdx: Proof_Stochastic_CD}. 
Compared with the deterministic bound, Theorem~\ref{thm_cd_sto} contains additional variance terms arising from stochastic UL gradient estimation and stochastic Jacobian estimation. 
The product term $\sigma_y^2\sigma_{f_y}^2$ reflects the interaction between the stochasticity in the estimated Jacobian and the stochasticity in the UL gradient with respect to $\y$.

In Theorems~\ref{thm_cd} and~\ref{thm_cd_sto}, the LL approximation and stochastic-error terms scale linearly with the UL dimension $d_u$, while the LL dimension $d_l$ enters the admissible step-size bound through the finite-difference error. 
Moreover, each full coordinate sweep requires $d_u+1$ LL-response queries, yielding $\mathcal{O}(d_uT)$ oracle calls over $T$ sweeps. 
Existing techniques for accelerating coordinate-wise methods can potentially be integrated into the \algcd~framework to further improve its efficiency.
We also discuss extensions of our convergence analysis to the case of set-valued, i.e., non-unique, LL solutions. 
Further details on both aspects are provided in Appendix~\ref{sec: Limitations}.

The following corollary gives the resulting stochastic convergence rate under a decreasing LL approximation error.

\begin{corollary}\label{cor: general_delta_sto}
    Under the same conditions as in Theorem~\ref{thm_cd_sto}, choose $\alpha= \Theta\left( T^{-1/2} \right)$ and suppose that the LL approximation error decreases geometrically, i.e., $\delta_{t+1}^2\le \left( 1-\gamma \right)\delta_{t}^2$ for some $0<\gamma<1$.
    Then, the iterates generated by \algcd~satisfy: $\frac{1}{T}  \sum_{t=0}^{T-1} \mathbb{E}\left[ \left\|\nabla \ell\left(\x_t\right)\right\|^2 \right] = \mathcal{O}\left( T^{-1/2} \right)$.
\end{corollary}

Corollary~\ref{cor: general_delta_sto} demonstrates that \algcd~achieves an $\mathcal{O}\left( T^{-1/2} \right)$ convergence rate for reaching a stationary solution. 
This rate matches the state-of-the-art stochastic convergence rate achieved by standard bilevel methods under stronger LL strong-convexity assumptions~\cite{ghadimi2018approximation, ji2021bilevel, chen2021tighter, arbel2021amortized, dagreou2022framework}.


\section{Numerical Experiments} \label{sec: NumericalResults}

In this section, we evaluate the performance of \algcd~on three different bilevel optimization tasks: 1) deep hyper-representation; 2) adversarial training; and 3) data weighting for LLM alignment.
The experiments are designed to examine two questions: whether \algcd~can achieve competitive performance under a weaker LL information model than existing bilevel methods, and how sensitive the proposed pseudo-hypergradient is to the quality of the approximate LL solution.
Due to space limitations, additional experimental details and results are provided in Appendix~\ref{apdx: Experiments}.

\begin{figure*}[t!]
\centering
\begin{minipage}[t]{0.235\textwidth}
    \centering
    \includegraphics[width=1\textwidth]{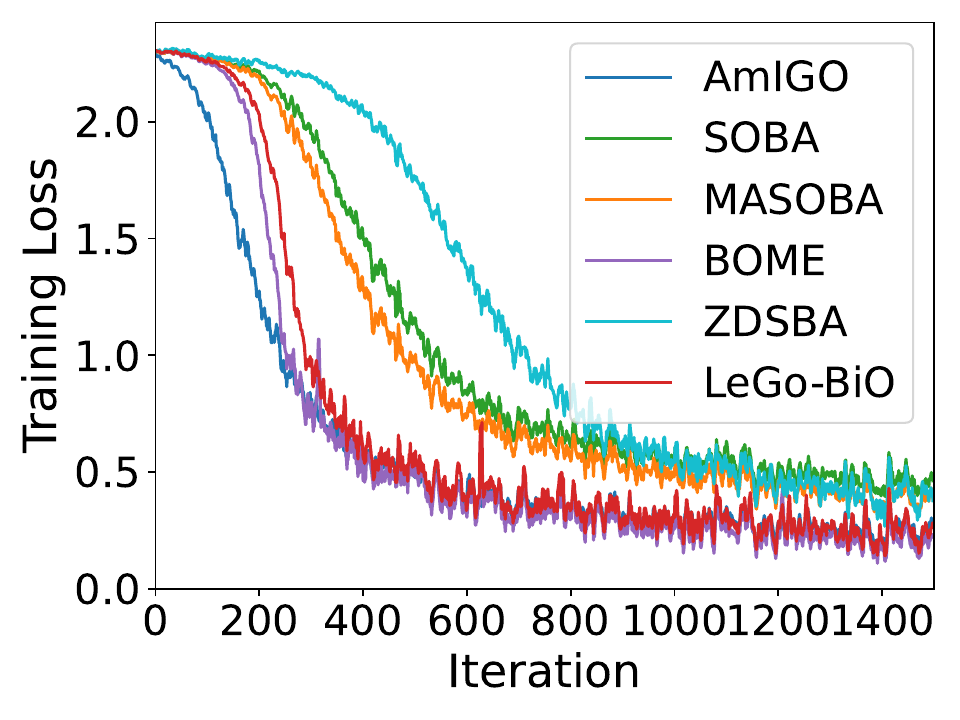}
    \caption{Performance among different algorithms on Task 1.}
    \label{fig:loss_convergence}
\end{minipage}
\hspace{.01\textwidth}
\begin{minipage}[t]{0.48\textwidth}
    \begin{subfigure}[t]{0.49\textwidth}
        \centering
        \includegraphics[width=1\textwidth]{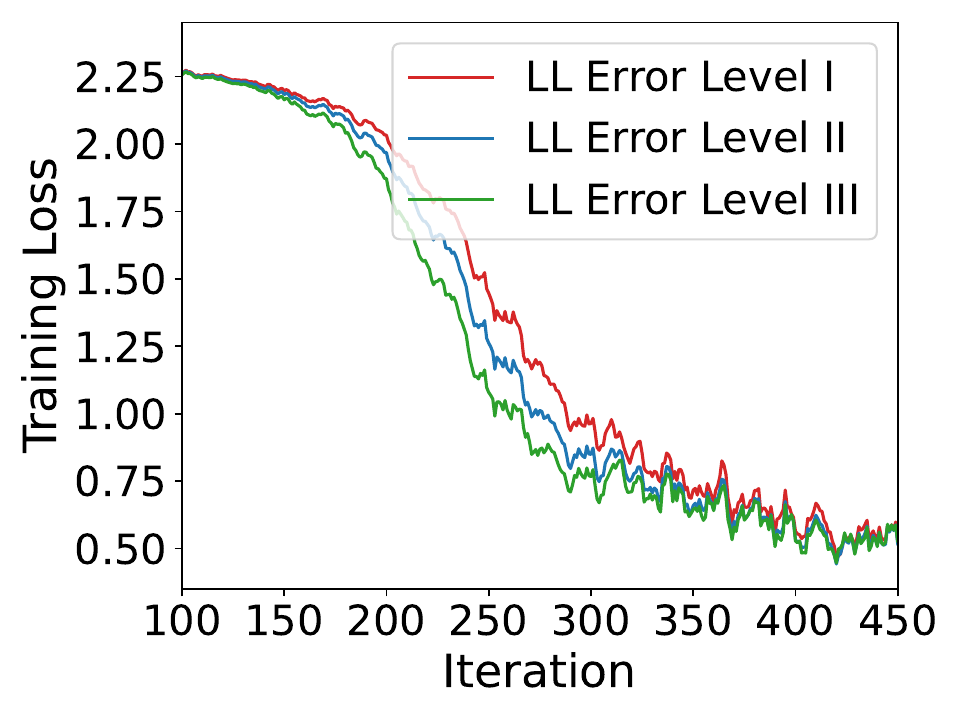}
        \caption{SGD}
    \end{subfigure}
    \begin{subfigure}[t]{0.49\textwidth}
        \centering
        \includegraphics[width=1\textwidth]{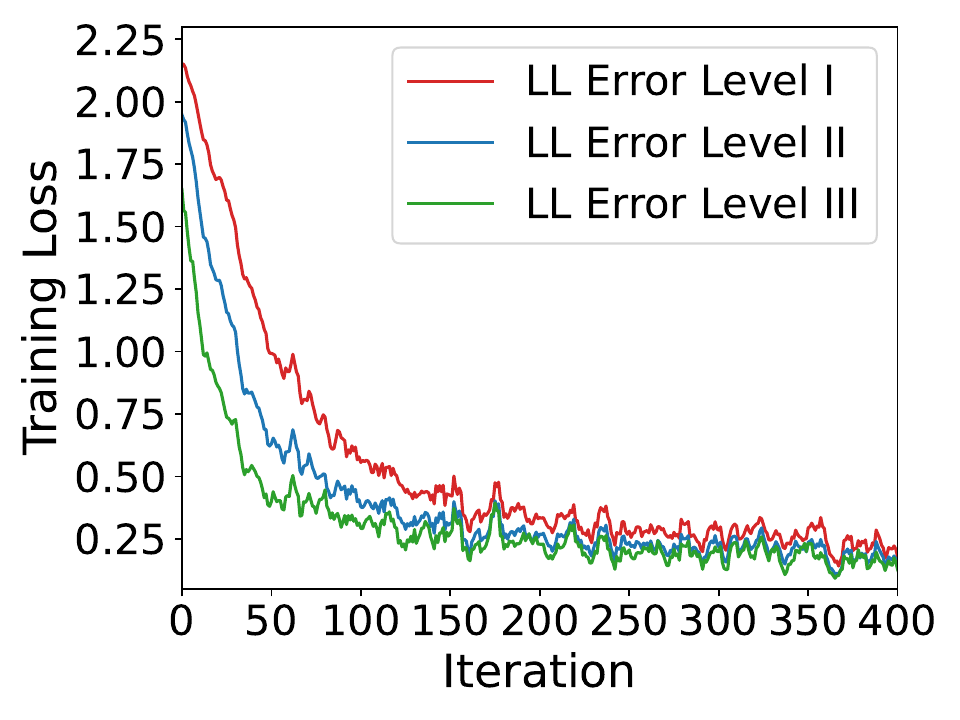}
        \caption{SciPy}
    \end{subfigure}
    \caption{Performance of \algcd~under different LL approximation errors on Task 1 (I is the largest).}  
    \label{fig:loss_error}
\end{minipage}
\hspace{.01\textwidth}
\begin{minipage}[t]{0.235\textwidth}
    \centering
    \includegraphics[width=1\textwidth]{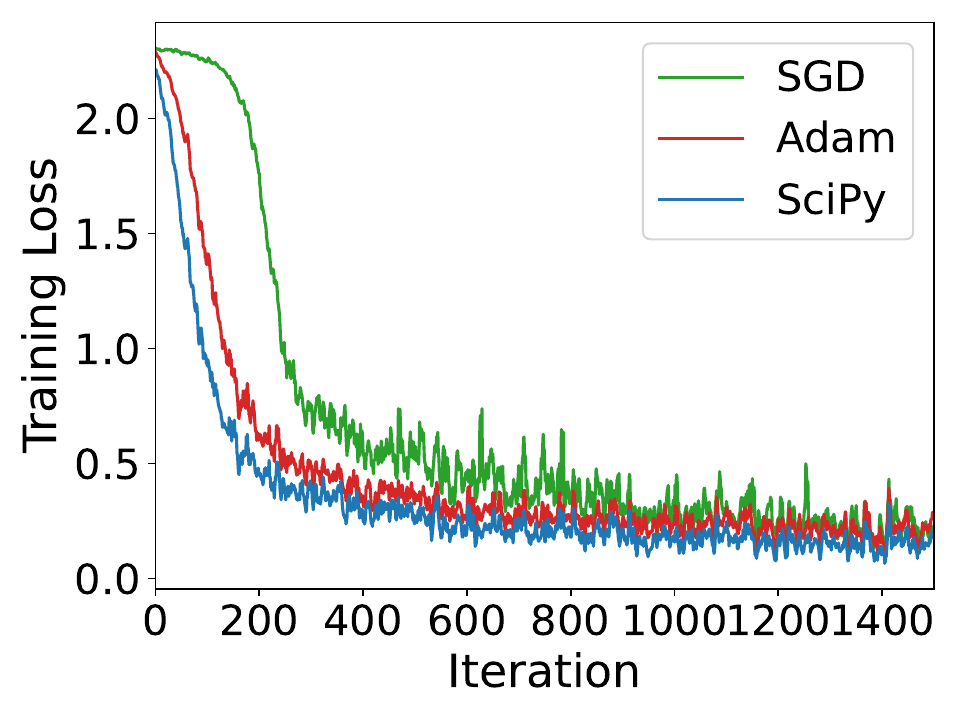}
    \caption{Performance of \algcd~under different LL solvers on Task 1.}
    \label{fig:loss_solver}
\end{minipage}
\vspace{-0.12in}
\end{figure*}

\paragraph{Task 1) Deep Hyper-Representation.}

We first evaluate \algcd~on a deep hyper-representation task, where the goal is to learn a feature representation such that the LL classifier induced by this representation performs well on image classification.
This task naturally reflects the LL-agnostic setting when the UL learner can observe the classifier returned by an external training procedure (an SGD-based classifier-training routine in our experiments), but does not have access to the LL training objective, its derivatives, or the procedure that generated the classifier.
We conduct experiments on the MNIST dataset using a 5-layer LeNet network~\cite{lecun1998gradient}.


We compare \algcd~with the AID-based methods AmIGO~\cite{arbel2021amortized}, SOBA~\cite{dagreou2022framework}, and MA-SOBA~\cite{chen2024optimal}, the fully first-order method BOME~\cite{liu2022bome}, and the zeroth-order method ZDSBA~\cite{aghasi2025fully}.
AmIGO, SOBA, and MA-SOBA use LL derivative information to approximate hypergradients, while BOME still requires LL first-order gradients; ZDSBA instead constructs zeroth-order estimates from function-value queries of the UL and LL objectives. 
In contrast, \algcd~uses only UL gradients and LL response queries.

\cref{fig:loss_convergence} compares the performance of different algorithms on the deep hyper-representation task. 
\algcd~achieves a training loss comparable to those of AmIGO and BOME and lower than those of SOBA, MA-SOBA, and ZDSBA.
These results demonstrate that, despite having no access to the LL objective or its derivative information, \algcd~can construct an effective pseudo-hypergradient using only UL gradients and LL response queries alone.
Moreover, the comparison with ZDSBA highlights the practical advantage of preserving the available UL first-order information rather than relying entirely on zeroth-order estimation.

We further study the robustness of \algcd~under different levels of LL solution approximation error. 
\cref{fig:loss_error} reports the performance of \algcd~under different LL approximation error levels, where Level I corresponds to the largest error. 
The results show that the final training losses remain comparable across different error levels.
At the same time, when the LL approximation error is smaller, \algcd~requires fewer iterations to reach the same training loss, due to a more accurate pseudo-hypergradient estimate.
These results demonstrate that \algcd~is robust to moderate LL approximation errors.
Moreover, these observations are consistent with our theoretical finding, where the convergence depends explicitly on the LL approximation error.

To evaluate whether \algcd~depends on a particular LL black-box solver, we have conducted a solver-diversity experiment using two simulated LL solvers, SGD and Adam, and an external SciPy-based solver.
As shown in~\cref{fig:loss_solver}, \algcd~remains effective across all three solvers, and SciPy achieves the lowest training loss. 
These results indicate that \algcd~can operate effectively with heterogeneous black-box response generators.

\paragraph{Task 2) Adversarial Training.}

\begin{wraptable}{r}{0.72\textwidth}
\vspace{-.16in}
\centering
\caption{Performance among different algorithms on Task 2.}
\label{table:adversarial}
\begin{small}
\setlength\tabcolsep{3.6pt}
\begin{tabular}{c|cccc}
\hline
                       & LeGo-BiO & FAST-BAT & FAST-AT & FAST-AT-GA \\ \hline
Standard Accuracy (\%) & 77.18    & 76.04    & 73.34   & 71.40      \\
Robust Accuracy (\%)   & 39.08    & 39.84    & 32.56   & 36.44      \\ \hline
\end{tabular}
\end{small}
\vspace{-0.1in}
\end{wraptable}

We next evaluate \algcd~on an adversarial training task, which aims to train a model that remains accurate under adversarial perturbations. 
In this bilevel formulation, the UL objective corresponds to the defender's model-training objective, while the LL problem corresponds to the attacker that returns adversarial perturbations against the current model.
This task is well aligned with the LL-agnostic setting when the defender observes the perturbations generated by an external or black-box attacker (implemented using projected gradient descent in our experiments), but cannot access the attack objective or differentiate through the attack-generation procedure.
We conduct experiments by training the PreActResNet-18~\cite{he2016identity} model on the CIFAR-10~\cite{krizhevsky2009learning} dataset.

We compare \algcd~with three fast adversarial training baselines: FAST-AT~\cite{wong2020Fast}, FAST-AT-GA~\cite{andriushchenko2020understanding}, and FAST-BAT~\cite{zhang2022revisiting}.
These baselines operate under a stronger information model, assuming that the attacker’s objective is known to the defender and require access to it for specification or differentiation during training.
In contrast, \algcd~only uses the perturbations returned by the attacker together with UL gradient information, and does not require access to the attack objective.

\cref{table:adversarial} reports standard accuracy on clean test images and robust accuracy under adversarial perturbations. 
Both \algcd~and FAST-BAT outperform FAST-AT and FAST-AT-GA in terms of both standard and robust accuracy, which supports the effectiveness of bilevel modeling for adversarial training.
Moreover, \algcd~achieves the highest standard accuracy among all compared methods, while attaining robust accuracy comparable to FAST-BAT and higher than FAST-AT and FAST-AT-GA. 
These results indicate that, despite not using the LL attack objective and its derivative information, \algcd~can construct an accurate pseudo-hypergradient that approximates the UL-gradient comparably to hypergradient-based methods, thereby achieving a favorable standard-robust accuracy trade-off under a weaker information model.

\paragraph{Task 3) Data Weighting for LLM Alignment.}

\begin{wrapfigure}{r}{0.29\textwidth}
\vspace{-.14in}
    \centering
    \includegraphics[width=1\linewidth]{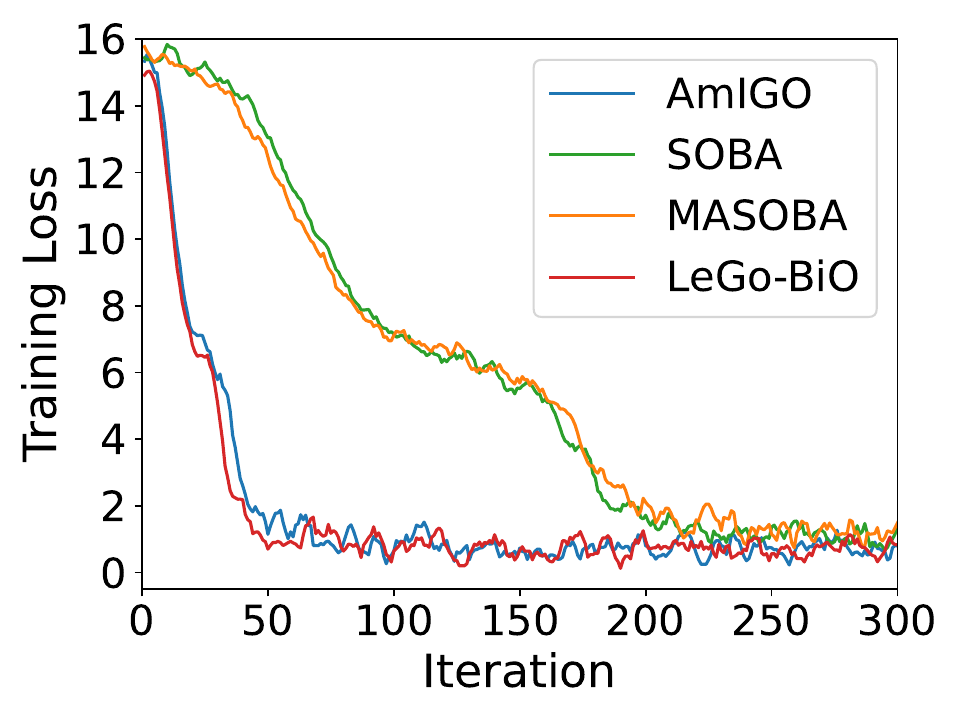}
    \caption{Performance among different algorithms on Task 3.}
    \label{fig:loss_convergence_dw}
\vspace{-.18in}
\end{wrapfigure}

We then evaluate \algcd~on a data-weighting task for LLM Alignment, with the goal of learning data weights for LLM alignment. 
Specifically, we align the Llama-3.2-1B-Instruct model~\cite{llama} on the HH-RLHF dataset~\cite{bai2022training}. 
For \algcd, LLM fine-tuning is treated as an external response oracle that maps data weights to fine-tuned LLM parameters; \algcd~neither accesses the LL training objective nor differentiates through the fine-tuning procedure.


We compare \algcd~with the white-box methods AmIGO~\cite{arbel2021amortized}, SOBA~\cite{dagreou2022framework}, and MA-SOBA~\cite{chen2024optimal}, which have full access to the LL objective and its derivatives.
This comparison therefore evaluates optimization performance under different information-access assumptions.

As shown in~\cref{fig:loss_convergence_dw}, \algcd~achieves a training loss comparable to those of AmIGO, SOBA and MA-SOBA. 
Moreover, \algcd~and AmIGO converge substantially faster than SOBA and MA-SOBA, likely because they obtain more accurate LL solutions. 
These results demonstrate that, despite having no access to the LL objective or its derivative information, \algcd~can construct an effective pseudo-hypergradient using only UL gradients and LL response queries.


\section{Conclusion} 

This paper studied lower-level (LL) agnostic bilevel optimization, where the upper-level (UL) learner observes only an approximate LL solution $\bar{\y}(\x)$ and has no access to the LL objective, its gradients, Hessian/Jacobian information, or the procedure that generates the response. 
We proposed \algcd, a coordinate-wise pseudo-hypergradient method that estimates the missing Jacobian $\nabla _{\x}\bar{\y}$ through finite differences of $\bar{\y}(\cdot)$ while retaining the available first-order gradients of the UL objective. 
The coordinate-wise construction isolates the Jacobian column required for each UL-gradient component, thereby avoiding the directional-projection limitation of full-parameter responses and the noise sensitivity of finite-differencing the entire UL-gradient.
We established convergence guarantees for nonconvex UL objectives in both deterministic and stochastic settings under an LL error-bound/Polyak–Łojasiewicz-type condition, without requiring LL strong convexity. 
Our results yield an $\mathcal{O}(T^{-1})$ deterministic rate and an $\mathcal{O}(T^{-1/2})$ stochastic rate when the LL approximation error decreases geometrically. 
Experiments on deep hyper-representation, adversarial training, and data weighting for LLM alignment show that \algcd~achieves competitive or superior performance despite having no access to the LL objective. 
These results suggest that LL response-based UL-gradient approximation offers a promising direction for bilevel optimization with black-box solvers, simulators, and learned LL modules.

\section*{Acknowledgements}

This work is supported in part by NSF grants CAREER CNS-2110259, CNS-2112471, ECCS-233104, DARPA YFA D24AP00265, DARPA HR0011-25-2-0019, ONR grant N00014-24-1-2729, and a Meta Research Award INB3267726.

\bibliographystyle{IEEEtran}{}
\bibliography{Reference}


\appendix


\section{Experimental Details and Additional Results} \label{apdx: Experiments}

In this section, we present additional experimental results and provide further details on the experimental setup and implementation.

\subsection{Additional Experimental Results}

\subsubsection{Deep Hyper-Representation}

\begin{figure*}[ht]
    \centering
    \includegraphics[width=0.4\textwidth]{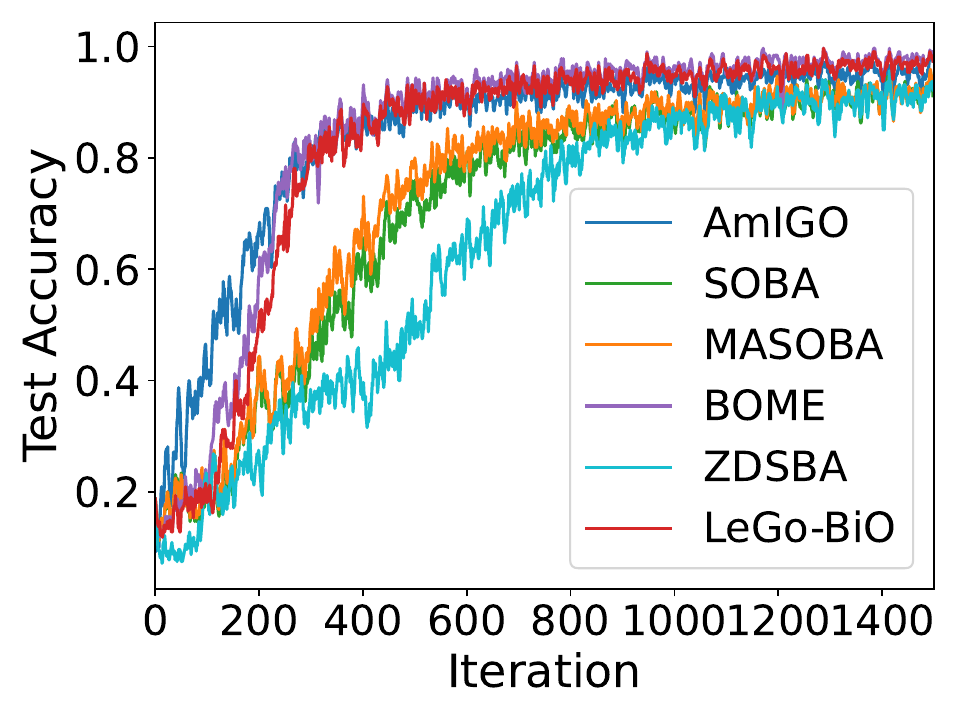}
    \caption{Comparison of test accuracy across different algorithms on Task 1.}
    \label{fig:acc_convergence}
\end{figure*}

\cref{fig:acc_convergence} compares the test accuracy of different algorithms on the deep hyper-representation task. 
\algcd~achieves test accuracy comparable to AmIGO and BOME, while consistently outperforming SOBA, MA-SOBA, and ZDSBA. 
Together with the training-loss results in~\cref{fig:loss_convergence}, these results demonstrate that \algcd~achieves competitive optimization and test performance despite having no access to the LL objective or its derivative information.
In particular, the comparison with ZDSBA further demonstrates the benefit of preserving the available UL first-order information instead of estimating the entire UL gradient through zeroth-order function-value queries.

\begin{figure*}[ht]
    \centering
    \begin{subfigure}[t]{0.4\textwidth}
        \centering
        \includegraphics[width=1\textwidth]{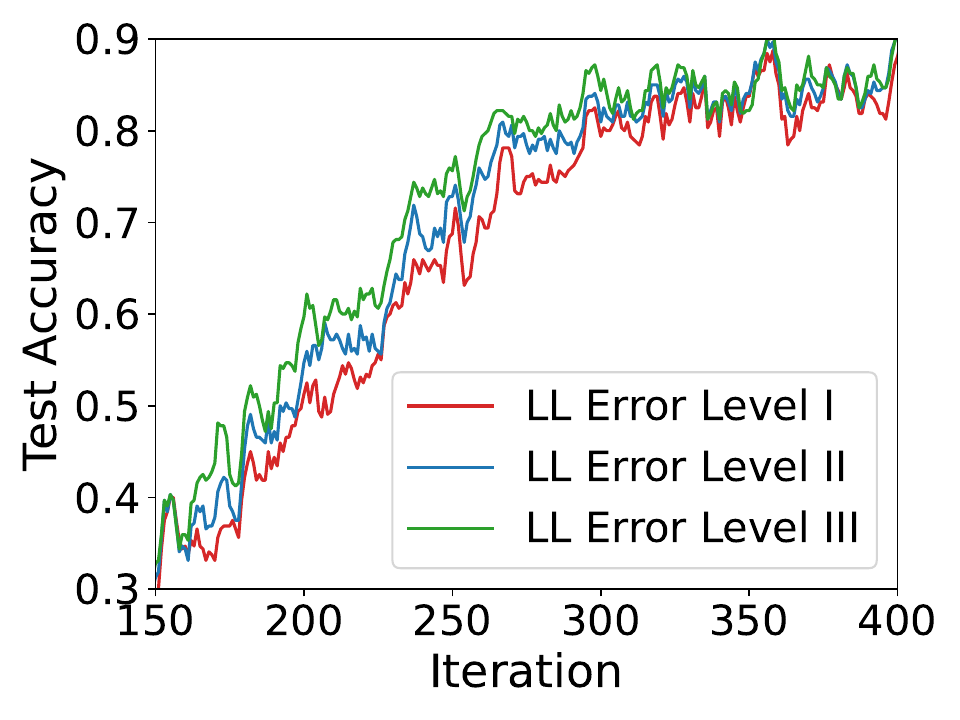}
        \caption{SGD}
    \end{subfigure}
    \centering
	\begin{subfigure}[t]{0.4\textwidth}
        \centering
        \includegraphics[width=1\textwidth]{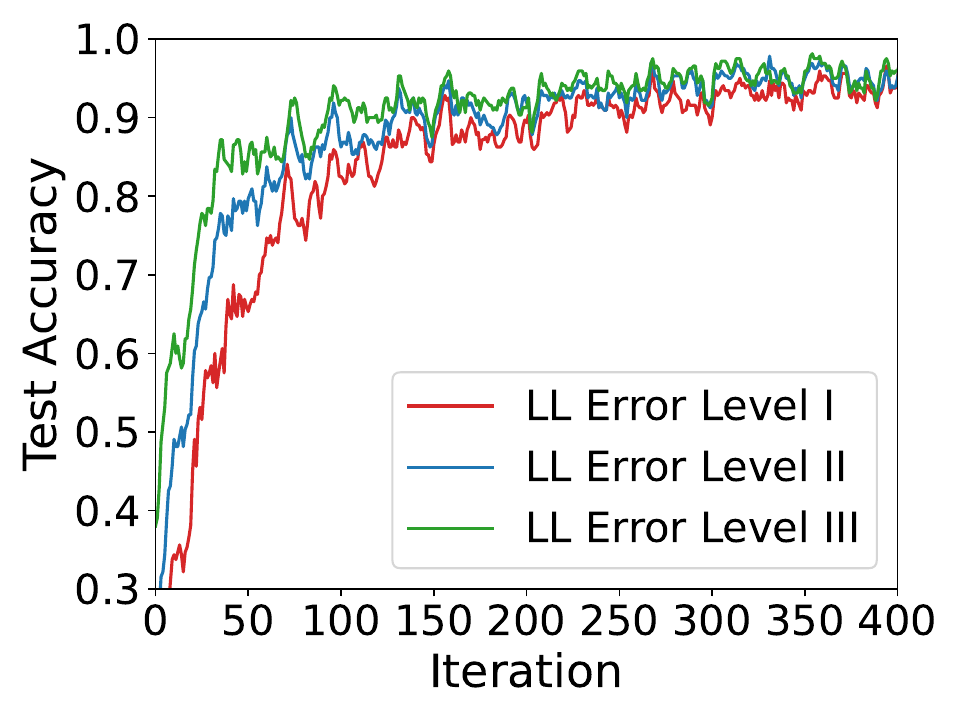}
        \caption{SciPy}
    \end{subfigure}
    \caption{Test accuracy of \algcd~under different LL approximation errors on Task 1 (I is the largest).}
    \label{fig:acc_error}
\end{figure*}

We further evaluate the effect of different LL approximation errors on the test accuracy of \algcd. 
As shown in~\cref{fig:acc_error}, \algcd~achieves similar final test accuracy across all three LL error levels for both the SGD-based and SciPy-based LL solvers. 
Smaller LL approximation errors generally lead to faster improvements in test accuracy, which is consistent with the training-loss results in~\cref{fig:loss_error}. 
These observations indicate that \algcd~is robust to moderate LL approximation errors, while more accurate LL responses can improve its convergence behavior.

\begin{figure*}[ht]
    \centering
    \includegraphics[width=0.4\textwidth]{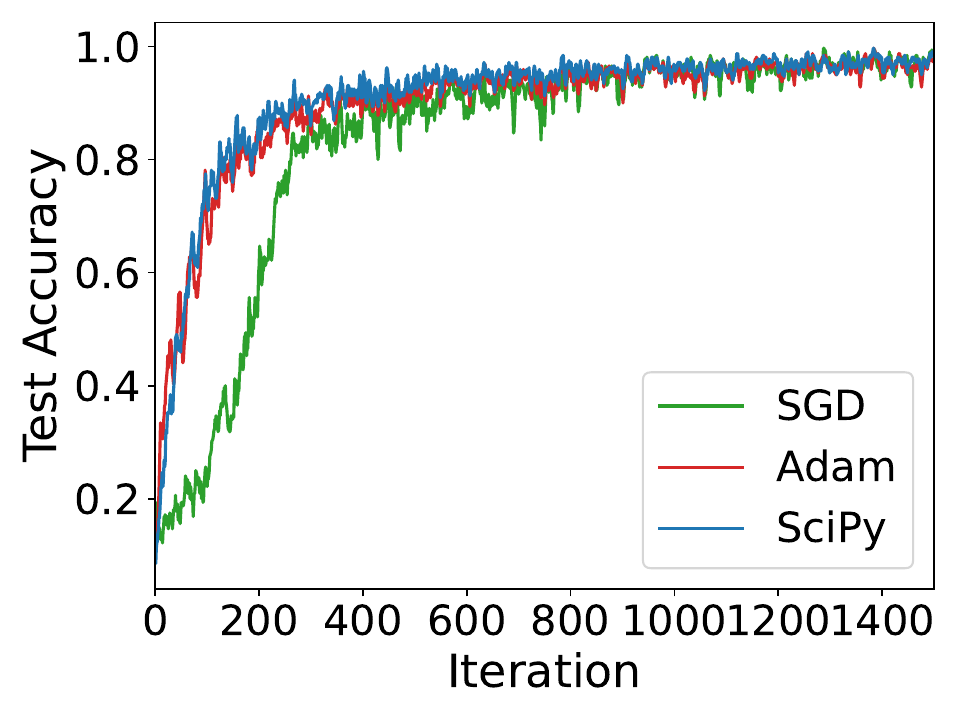}
    \caption{Test accuracy of \algcd~under different LL solvers on Task 1.}
    \label{fig:acc_solver}
\end{figure*}

Finally, \cref{fig:acc_solver} reports the test accuracy of \algcd~under different LL solvers. 
\algcd~achieves comparable final test accuracy when the LL responses are generated by SGD, Adam, or SciPy, although their transient convergence behaviors differ. 
This result complements the training-loss comparison in~\cref{fig:loss_solver} and further shows that \algcd~does not rely on a particular LL solver. 
Instead, it can effectively utilize approximate responses produced by heterogeneous black-box LL procedures.

\subsection{Experimental Details}

\subsubsection{Deep Hyper-Representation}

Deep hyper-representation learns a feature extractor by optimizing the validation performance of a classifier trained on the induced features. 
Let $\lambda$ denote the parameters of the feature extractor $\phi_\lambda$ and let $w$ denote the classifier parameters. 
The bilevel objective is
\begin{align*}
    &\min_{\lambda}~ f(\lambda,w^\star(\lambda))
    := \frac{1}{|\mathcal{D}_{\rm val}|}
    \sum_{(a_i,b_i)\in\mathcal{D}_{\rm val}}
    \ell_{\rm ce}\big(w^\star(\lambda);\phi_\lambda(a_i),b_i\big),\\
    &\text{s.t.}~~
    w^\star(\lambda) = \arg\min_w~g(\lambda,w)
    := \frac{1}{|\mathcal{D}_{\rm tr}|}
    \sum_{(a_i,b_i)\in\mathcal{D}_{\rm tr}}
    \ell_{\rm ce}\big(w;\phi_\lambda(a_i),b_i\big),
\end{align*}
where $\left(a_i, b_i\right)$ denotes a data sample, $\mathcal{D}_{val}$ and $\mathcal{D}_{tr}$ are the validation dataset and the training dataset, $\ell_{\rm ce}$ corresponds to the cross-entropy loss.

In the LL-agnostic setting, \algcd~does not access $g(\lambda,w)$ or differentiate through the LL classifier-training routine. 
For each queried representation parameter $\lambda$, the algorithm only receives an approximate classifier $\bar{w}(\lambda)$, which is produced by an external SGD-based classifier-training routine in our experiments, and uses coordinate-wise response differences of $\bar{w}(\cdot)$ to estimate the missing Jacobian contribution in the UL gradient.

We conduct experiments using a LeNet-5 network~\cite{lecun1998gradient} (MIT License) on the MNIST dataset~\cite{lecun1998gradient} (CC BY-SA 3.0 License) with batch size $64$.
Here, the UL variable $\mathbf{x}=\lambda$ has dimension 60,856, while the LL variable $\mathbf{y}=w$ has dimension 850.
All methods use the same data preprocessing, model architecture, and train/test split. 
We compare \algcd~against AmIGO~\cite{arbel2021amortized}, SOBA~\cite{dagreou2022framework}, and MA-SOBA~\cite{chen2024optimal}.
For \algcd, we set the step size to $\alpha=0.01$ and simulate the LL solver using stochastic gradient descent with step size $\beta=0.001$.
For both AmIGO and MA-SOBA, we set the step sizes to $\alpha=0.01$, $\beta=0.1$, and $\gamma=0.1$.
For AmIGO, we use $8$ $\y$-update iterations and $4$ $\z$-update iterations.
For MA-SOBA, we set the momentum coefficient to $0.9$. 
For SOBA, we set the step sizes to $\alpha=0.01$, $\beta=0.1$, and $\gamma=0.01$. 
All deep hyper-representation experiments are conducted on an NVIDIA H200 NVL GPU.

\subsubsection{Adversarial Training}

We formulate adversarial training as a bilevel optimization problem, where the UL variable is the classifier parameter $\theta$ and the LL response is the adversarial perturbation. 
For each data-label pair $(a,b)$, the LL attack problem is
\begin{align}
    \delta^\star(\theta) \in \arg\max_{\delta\in\mathcal{C}_{\epsilon}(a)}
    \ell_{\rm ce}(\theta;a+\delta,b),
    \quad
    \mathcal{C}_{\epsilon}(a) := \{\delta: \|\delta\|_{\infty}\leq \varepsilon,~a+\delta\in[0,1]^d\},
    \label{eq:adv_ll_apdx}
\end{align}
where $\varepsilon$ is the perturbation budget and $\ell_{\rm ce}$ is the cross-entropy loss. The UL objective is the empirical robust training loss
\begin{align*}
    \min_{\theta}~
    \frac{1}{|\mathcal{D}_{\rm tr}|}
    \sum_{(a_i,b_i)\in\mathcal{D}_{\rm tr}}
    \ell_{\rm ce}\big(\theta;a_i+\delta_i^\star(\theta),b_i\big).
\end{align*}
Although Eq.~(\ref{eq:adv_ll_apdx}) defines the target LL response, \algcd~does not use the LL attack objective or its derivatives. 
Instead, the attack routine is treated as an external response oracle: for a queried classifier parameter $\theta$, the oracle returns an approximate perturbation $\bar{\delta}(\theta)$ based on projected gradient descent. 
The returned perturbation is detached from the attack-generation procedure, and \algcd~uses finite differences of the returned perturbations to estimate the missing Jacobian information.

We conduct experiments using the PreActResNet-18~\cite{he2016identity} model (MIT License) on the CIFAR-10 dataset~\cite{krizhevsky2009learning} (CC BY 4.0 License) with batch size $128$.
Here, the UL variable $\mathbf{x}=\theta$ has dimension 11,172,170, while the LL variable $\mathbf{y}=\delta$ has dimension 3,072.
The perturbation set is the $\ell_\infty$ ball with training and evaluation budget $\varepsilon=10/255$.
All methods use the same data preprocessing, model architecture, and train/test split. 
For standard accuracy, we evaluate the trained classifier on clean test images. 
For robust accuracy, we evaluate the classifier on adversarially perturbed test images generated under the same $\ell_\infty$ budget.
Specifically, we use PGD-50-10, i.e., a $50$-step projected gradient descent attack with $10$ random restarts.

We compare \algcd~against FAST-AT~\cite{wong2020Fast}, FAST-AT-GA~\cite{andriushchenko2020understanding}, and FAST-BAT~\cite{zhang2022revisiting}. 
For FAST-AT, we use a constant step size of $0.1$. For both FAST-AT-GA and FAST-BAT, we use a constant step size of $0.05$. For \algcd, we use a cyclic step size ranging from $0$ to $2.0$.
All the experiments are conducted on an NVIDIA H200 NVL GPU.

\subsection{Data Weighting for LLM Alignment}

In this subsection, we describe the experimental setup for the data weighting task for LLM alignment.
The goal is to learn a set of training-data weights such that fine-tuning the LLM on the reweighted training data yields low validation loss.

We partition the training data into $N_T$ groups, denoted by $\mathcal{T}_n$, where $n=1,2,\ldots,N_T$.
Each data sample $i=1,2,\ldots,\vert\mathcal{T}_n\vert$ consists of a prompt-response pair $\{p_{n,i},r_{n,i}\}$ and its corresponding labeled score $s_{n,i}$.
We introduce $\mathbf{x}=[x_1,x_2,\ldots,x_{N_T}]^\top$ as a vector of raw dataset weights and convert it into normalized weights through a softmax transformation.
Let $\mathbf{y}\in\mathbb{R}^l$ denote the trainable parameters of the LLM.
The UL problem optimizes the dataset weights based on the loss evaluated on a validation set $\mathcal{V}$, while the LL problem fine-tunes the LLM using the weighted training data.

Accordingly, the bilevel optimization problem is formulated as
\begin{align*}
&\min_{\mathbf{x}\in\mathbb{R}^{N_T}}\sum_{i=1}^{\vert\mathcal{V}\vert} \mathcal{L}(\tilde{r}_{0,i},r_{0,i}; p_{0,i},\mathbf{y}^*(\mathbf{x})) \\
&\text{s.t.}\quad \mathbf{y}^*(\mathbf{x})=\argmin_{\mathbf{y}\in\mathbb{R}^l}\sum_{n=1}^{N_T}\frac{e^{x_n}}{\sum_{n'=1}^{N_T} e^{x_{n'}}}\sum_{i=1}^{\vert\mathcal{T}_n\vert} \mathcal{L}(\tilde{r}_{n,i},r_{n,i}; p_{n,i},\mathbf{y}) \nonumber,
\end{align*}
where $\mathcal{L}(\tilde{r}_{n,i},r_{n,i}; p_{n,i},\mathbf{y})$ is the loss between the true response $r_{n,i}$ and the response $\tilde{r}_{n,i}$ generated by the LLM parameterized by $\mathbf{y}$ for prompt $p_{n,i}$.

We use Llama-3.2-1B-Instruct~\cite{llama} (Llama 3.2 Community License) as the base model and fine-tune it using the low-rank adaptation (LoRA) technique with rank $8$. 
Experiments are conducted on the Anthropic HH-RLHF dataset~\cite{bai2022training} (MIT License).
The training samples are divided into two groups according to their preference labels, namely \textit{chosen} and \textit{rejected}, so that $N_T=2$. 
The validation set contains only \textit{chosen} responses.
Consequently, an effective data-weighting strategy is expected to assign relatively larger weight to the \textit{chosen} training group, thereby reducing the validation loss. 
Cross-entropy loss is used for both the UL and LL objectives.

For AmIGO~\cite{arbel2021amortized}, we use $5$ update steps for both $\mathbf{y}$ and $\mathbf{z}$.
For MA-SOBA~\cite{chen2024optimal}, we set $\mu=0.8$.
All methods use step sizes $\alpha=5\times10^{-3}$, $\beta=2\times10^{-4}$, and $\gamma=3\times10^{-4}$, with a batch size of $32$. 
All experiments are conducted on an NVIDIA H200 NVL GPU.

\section{Discussion} \label{sec: Limitations}

\subsection{Coordinate-Wise Design}


Due to its coordinate-wise design, a full \algcd~sweep requires $\mathcal{O}(d_u)$ LL-response queries, and this dimension dependence also appears in our convergence bounds. 

This cost reflects an explicit "exploitation–exploration dilemma" in our LL-agnostic setting, which shows that the black-box bilevel optimization problem is {\em fundamentally hard}. 
A single full-parameter response difference provides only one directional projection of the Jacobian and therefore cannot recover the full pseudo-hypergradient without additional perturbations/explorations or structural assumptions. 
The coordinate-wise queries instead isolate each component, avoiding this projection ambiguity while preserving the available UL first-order information.

Consequently, the current full-sweep method is best suited to problems with a moderate effective UL dimension, or settings in which $\mathbf{x}$ has a structured low-dimensional parameterization, such as hyperparameters, layer-wise coefficients, or adapter parameters. 
The LL queries can also be batched or evaluated in parallel (if given an appropriate distributed computing environment) once the query points are formed, reducing the LL response latency.

For large networks with an appropriate symmetry, a natural scalable extension is to query only $k\ll d_u$ randomly selected coordinates per iteration, reducing the per-iteration LL-query cost to $\mathcal{O}(k)$. 
Another option is to apply \algcd~to a structured low-dimensional reparameterization or develop block-wise or layer-wise estimators. 
These variants introduce additional stochastic or projection error and require a separate convergence analysis, which we identify as an important direction for future work.

\subsection{Extension to Set-Valued LL Solutions}

It is possible to extend our analysis to the {\em set-valued LL solution setting} (i.e., {\em non-unique} LL solutions) under a bounded diameter condition. 
Specifically, assume the LL solution set has bounded dimension $D$, and the oracle selects a differentiable branch $\mathbf{y}^{\dagger}(\mathbf{x})$ satisfying $\left\| \bar{\mathbf{y}}\left(\mathbf{x}_t\right)-\mathbf{y}^{\dagger}(\mathbf{x}_t) \right\|\le C_D\delta_t$, where $C_D$ depends on the solution-set geometry, together with the corresponding Jacobian regularity of $\mathbf{y}^{\dagger}(\mathbf{x})$. 
Replacing $\mathbf{y}^*(\mathbf{x})$ with $\mathbf{y}^{\dagger}(\mathbf{x})$ and Assumption~\ref{asmp_y}(1) accordingly, the existing proof will carry over with the LL approximation-error terms scaled by $C_D^2$.


\section{Supporting Lemmas}

\begin{lemma}\label{lemma_Lip}
     Under Assumptions~\ref{asmp_up} and \ref{asmp_low}, we have $\left\|\nabla \ell\left(\x_{1}\right)-\nabla \ell\left(\x_{2}\right)\right\| \leq L_{l}\left\|\x_{1}-\x_{2}\right\|$, for all $\x_{1},\x_{2}\in \mathbb{R}^{d_u}$, where the Lipschitz constant is defined as $L_{l}=L_{f}+\kappa L_{f} L_{g_y}$.
\end{lemma}

\begin{proof}
We have
\begin{align} \label{eq:lip_l}
    &\left\|\nabla \ell\left(\x_{1}\right)-\nabla \ell\left(\x_{2}\right)\right\| \nonumber\\
    &=\left\|\nabla f\left(\x_{1}, \y^*\left( \x_1 \right)\right)-\nabla f\left(\x_{2}, \y^*\left( \x_2 \right)\right)\right\|\nonumber\\
    & \stackrel{(a)}{\le} L_{f}\left\|\x_1-\x_2\right\|+L_{f}\left\|\y^*\left( \x_1 \right)-\y^*\left( \x_2 \right)\right\|,
\end{align}
where (a) uses the Lipschitzness of $\nabla f\left(\x, \y\right)$ (see Assumption~\ref{asmp_up}).

Then, we bound $\left\|\y^*\left( \x_1 \right)-\y^*\left( \x_2 \right)\right\|$ in Eq.~(\ref{eq:lip_l}).
\begin{align}\label{eq:lip_y}
    &\left\|\y^*\left( \x_1 \right)-\y^*\left( \x_2 \right)\right\|\nonumber\\
    & \stackrel{(a)}{\le} \kappa\left\| \nabla _\y g\left( \x_1,\y^*\left( \x_2 \right)\right) \right\| \nonumber\\
    & \stackrel{(b)}{=} \kappa\left\| \nabla _\y g\left( \x_1,\y^*\left( \x_2 \right)\right)-\nabla _\y g\left( \x_2,\y^*\left( \x_2 \right)\right) \right\| \nonumber\\
    & \stackrel{(c)}{\le} \kappa L_{g_y}\left\| \x_1-\x_2 \right\|,
\end{align}
where (a) and (c) are because of Assumption~\ref{asmp_low}. (b) follows from $\nabla _\y g\left( \x_2,\y^*\left( \x_2 \right)\right)=0$.

Substituting Eq.~(\ref{eq:lip_y}) into Eq.~(\ref{eq:lip_l}) completes the proof.
\end{proof}

\begin{lemma}\label{lemma_zero_cd}
    For any function $\varphi(x): \mathbb{R} \to \mathbb{R}$, define one-direction zeroth-order estimator as $\widehat{\nabla} \varphi\left( x_1 \right) =\frac{1}{h}\left( \varphi\left( x_2 \right)-\varphi\left( x_1 \right) \right)$, where $h=x_2-x_1$. Assuming $\nabla \varphi\left( x \right)$ is Lipschitz with $L_\varphi\ge 0$, we have
    \begin{gather*}
        \left\| \widehat{\nabla} \varphi\left( x_1 \right)-\nabla \varphi\left( x_1 \right) \right\|^2\le \frac{h^2 L_\varphi^2}{4},
    \end{gather*}
    for all $x_{1}\in \mathbb{R}$.
\end{lemma}

\begin{proof}
For $h\neq 0$, we have
\begin{align*}
    &\left\| \widehat{\nabla} \varphi\left( x_1 \right)-\nabla \varphi\left( x_1 \right) \right\| \\
    & = \left\| \frac{1}{h}\left[ \varphi\left( x_1+h \right)-\varphi\left( x_1 \right) \right]-\nabla \varphi\left( x_1 \right) \right\|\\
    & \stackrel{(a)}{= } \left\| \int_0^1\big(\nabla\varphi(x_1+\rho h)-\nabla\varphi(x_1)\big)d\rho \right\|\\
    & \stackrel{(b)}{\le  } \int_0^1 L_\varphi\rho|h|d\rho\\
    &= \frac{1}{2}|h|L_\varphi,
\end{align*}
where (a) follows from the fundamental theorem of calculus. (b) utilizes the assumption that $\nabla \varphi\left( x \right)$ is Lipschitz.

Squaring proves the result.
\end{proof}

\begin{lemma} \label{lemma_q}
    Define $q_t^j= \frac{\bar{\y}(\x_{t-1})-\bar{\y}(\widetilde{\x}_t^j)}{x_{t-1}^j-x_t^j}$ and $\widehat{q}_t^j= \frac{\bar{\y}(\x_{t-1};\mathcal{B}^{y})-\bar{\y}(\widetilde{\x}_t^j;\mathcal{B}^{y})}{x_{t-1}^j-x_t^j}$. Under Assumptions~\ref{asmp_y} and ~\ref{asmp_sg}, we have
    \begin{gather*}
        \left\| q_t^j \right\|\le B_y, \qquad \mathbb{E}\left[ \left\| \widehat{q}_t^j-q_t^j \right\|^2 \right]\le \sigma_y^2.
    \end{gather*}

\end{lemma}

\begin{proof}
Apply the mean-value theorem along the line segment connecting $\widetilde{\x}_t^j$ and $\x_{t-1}$.

Define the line segment 
\begin{equation*}
    \nu_{t,j}\left( \rho \right)=\widetilde{\x}_t^j+\rho\left( \x_{t-1}-\widetilde{\x}_t^j \right),\qquad \rho\in \left[ 0,1 \right].
\end{equation*}

Since $\x_{t-1}-\widetilde{\x}_t^j =\left( x_{t-1}^j-x_t^j \right) \e_j$, we have
\begin{gather*}
    \nu_{t,j}\left( 0 \right)=\widetilde{\x}_t^j, \qquad \nu_{t,j}\left( 1 \right)=\x_{t-1}, \qquad \nu_{t,j}'\left( \rho \right)=\left( x_{t-1}^j-x_t^j \right) \e_j.
\end{gather*}

We first prove $\left\| q_t^j \right\|\le B_y$.

By the fundamental theorem of calculus,
\begin{equation*}
    \bar{\y}(\x_{t-1})-\bar{\y}(\widetilde{\x}_t^j)=\int_{0}^{1}\frac{d }{d\rho}\bar{\y}(\nu_{t,j}\left( \rho \right))d\rho.
\end{equation*}

Using the chain rule,
\begin{equation*}
    \frac{d }{d\rho}\bar{\y}(\nu_{t,j}\left( \rho \right))=\nabla _{\x}\bar{\y}(\nu_{t,j}\left( \rho \right))\nu_{t,j}'\left( \rho \right)=\nabla _{\x}\bar{\y}(\nu_{t,j}\left( \rho \right))\left( x_{t-1}^j-x_t^j \right) \e_j.
\end{equation*}

Therefore,
\begin{equation*}
    \bar{\y}(\x_{t-1})-\bar{\y}(\widetilde{\x}_t^j)=\int_{0}^{1}\nabla _{\x}\bar{\y}(\nu_{t,j}\left( \rho \right))\left( x_{t-1}^j-x_t^j \right) \e_jd\rho.
\end{equation*}

Divide both sides by $x_{t-1}^j-x_t^j$, we obtain
\begin{gather*}
    q_t^j=\int_{0}^{1}\nabla _{\x}\bar{\y}(\nu_{t,j}\left( \rho \right))\e_jd\rho.
\end{gather*}

Now take the norm:
\begin{align*}
    &\left\| q_t^j \right\|=\left\| \int_{0}^{1}\nabla _{\x}\bar{\y}(\nu_{t,j}\left( \rho \right))\e_jd\rho \right\|\\
    &\stackrel{(a)}{\le } \int_{0}^{1}\left\| \nabla _{\x}\bar{\y}(\nu_{t,j}\left( \rho \right))\e_j \right\|d\rho\\
    &\le \int_{0}^{1}\left\| \nabla _{\x}\bar{\y}(\nu_{t,j}\left( \rho \right)) \right\|\left\| \e_j \right\|d\rho\\
    &\stackrel{(b)}{\le }\int_{0}^{1}B_yd\rho\\
    &=B_y,
\end{align*}
where (a) utilizes the triangle inequality for integrals. (b) is because of Assumption~\ref{asmp_y} and $\left\| \e_j \right\|=1$.

Then, we prove $\mathbb{E}\left[ \left\| \widehat{q}_t^j-q_t^j \right\|^2 \right]\le \sigma_y^2$.

By the fundamental theorem of calculus,
\begin{equation*}
    \bar{\y}(\x_{t-1};\mathcal{B}^{y})-\bar{\y}(\widetilde{\x}_t^j;\mathcal{B}^{y})=\int_{0}^{1}\nabla _{\x}\bar{\y}(\nu_{t,j}\left( \rho \right);\mathcal{B}^{y})\left( x_{t-1}^j-x_t^j \right) \e_jd\rho.
\end{equation*}

Dividing by $x_{t-1}^j-x_t^j$, we have
\begin{gather*}
    \widehat{q}_t^j=\int_{0}^{1}\nabla _{\x}\bar{\y}(\nu_{t,j}\left( \rho \right);\mathcal{B}^{y})\e_jd\rho.
\end{gather*}

Hence,
\begin{equation*}
    \widehat{q}_t^j-q_t^j=\int_{0}^{1}\left( \nabla _{\x}\bar{\y}(\nu_{t,j}\left( \rho \right);\mathcal{B}^{y})-\nabla _{\x}\bar{\y}(\nu_{t,j}\left( \rho \right)) \right)\e_jd\rho.
\end{equation*}

Using Jensen’s inequality, we get
\begin{align*}
    &\mathbb{E}\left[ \left\| \widehat{q}_t^j-q_t^j \right\|^2 \right]\le \int_{0}^{1}\mathbb{E}\left[ \left\| \left( \nabla _{\x}\bar{\y}(\nu_{t,j}\left( \rho \right);\mathcal{B}^{y})-\nabla _{\x}\bar{\y}(\nu_{t,j}\left( \rho \right)) \right)\e_j \right\|^2 \right]d\rho\\
    &\stackrel{(a)}{\le }\int_{0}^{1}\sigma_y^2d\rho\\
    &=\sigma_y^2,
\end{align*}
where (a) uses Assumption~\ref{asmp_sg}.

These complete the proof.
\end{proof}

\section{Theoretical Analysis for \algcd~in Deterministic Setting} \label{apdx: Proof_Deterministic_CD}

Define the coordinate-wise estimator vector
\begin{equation*}
    \mathbf{d}_t^f \triangleq \sum_{j=1}^{d_u}\left[h_t^{f, j}\right]_j \e_j\in \mathbb{R}^{d_u}.
\end{equation*}

The update of the UL parameter can be written as
\begin{equation} \label{eq:x_update}
    \x_{t+1}=\x_{t}-\alpha_{t}\mathbf{d}_t^f.
\end{equation}

\subsection{Descent in the Upper-Level Objective Function}

\begin{lemma}\label{apx:lemma_l_cd}
     Under Assumptions~\ref{asmp_up} and \ref{asmp_low} and choosing $\alpha_t\le \frac{1}{2L_l}$, the following inequality holds for successive iterations of Algorithm~\ref{alg_cd}:
     \begin{align*}
        &\ell\left(\x_{t+1}\right)-\ell\left(\x_t\right) \le -\frac{\alpha_t}{2}\left\|\nabla \ell\left(\x_t\right)\right\|^2-\frac{\alpha_t}{4}\left\|\mathbf{d}_{t}^f\right\|^2+\frac{\alpha_t}{2}\left\|\nabla \ell\left(\x_t\right)-\mathbf{d}_{t}^f\right\|^2,
     \end{align*}
     for all $t\in \left\{ 0,1,\ldots,T-1  \right\}$.
\end{lemma}

\begin{proof}
We have
\begin{align*}
    &\ell\left(\x_{t+1}\right)-\ell\left(\x_t\right) \nonumber\\
    &\stackrel{(a)}{\leq} \left\langle\nabla \ell\left(\x_t\right), \x_{t+1}-\x_t\right\rangle+\frac{L_l}{2}\left\|\x_{t+1}-\x_t\right\|^2\nonumber\\
    &\stackrel{(b)}{=}-\alpha_t\left\langle\nabla \ell\left(\x_t\right), \mathbf{d}_{t}^f\right\rangle+\frac{\alpha_t^2 L_l}{2} \left\|\mathbf{d}_{t}^f\right\|^2\nonumber\\
    &\stackrel{(c)}{=} -\frac{\alpha_t}{2}\left\|\nabla \ell\left(\x_t\right)\right\|^2-\frac{\alpha_t}{2}\left\|\mathbf{d}_{t}^f\right\|^2+\frac{\alpha_t}{2}\left\|\nabla \ell\left(\x_t\right)-\mathbf{d}_{t}^f\right\|^2+\frac{\alpha_t^2 L_l}{2} \left\|\mathbf{d}_{t}^f\right\|^2 \nonumber \\
    &\stackrel{(d)}{\le} -\frac{\alpha_t}{2}\left\|\nabla \ell\left(\x_t\right)\right\|^2-\frac{\alpha_t}{4}\left\|\mathbf{d}_{t}^f\right\|^2+\frac{\alpha_t}{2}\left\|\nabla \ell\left(\x_t\right)-\mathbf{d}_{t}^f\right\|^2,
\end{align*}
where (a) uses the Lipschitz continuous gradients of $\ell$ (see Lemma~\ref{lemma_Lip}). (b) follows from the update rule in Eq.~(\ref{eq:x_update}). (c) is because of $\left\langle x,y \right\rangle=\frac{1}{2}\left\| x \right\|^{2}+\frac{1}{2}\left\| y \right\|^{2}-\frac{1}{2}\left\| x-y \right\|^{2}$. (d) chooses $\alpha_t\le \frac{1}{2L_l}$.

The proof of the lemma is complete.
\end{proof}

\subsection{Descent in the Gradient Approximation Error}
\begin{lemma}\label{apx:lemma_h_cd}
     Under Assumptions~\ref{asmp_up}--\ref{asmp_y}, the gradient approximation error in Algorithm~\ref{alg_cd} satisfies the following inequality:
     \begin{align*}
        &\left\|\nabla \ell\left(\x_t\right)-\mathbf{d}_{t}^f\right\|^2 \le 4d_uL_l^2\alpha_{t-1}^2\left\| \mathbf{d}_{t-1}^f \right\|^2+B_{f_y}^2d_l\frac{L_y^2}{2}\alpha_{t-1}^2\left\| \mathbf{d}_{t-1}^f \right\|^2+4d_uL_f^2\delta_t^2,
    \end{align*}
    for all $t\in \left\{ 0,1,\ldots,T-1  \right\}$.
\end{lemma}

\begin{proof}
We have
\begin{align}\label{eq:error_l_cd}
    &\left\|\nabla \ell\left(\x_t\right)-\mathbf{d}_{t}^f\right\|^2 \nonumber\\
    &= \sum_{j=1}^{d_u}\left( \left[ \nabla \ell\left(\x_t\right) \right]_j-\left[ h_{t}^{f,j} \right]_j \right)^2 \nonumber\\
    &\le  2\sum_{j=1}^{d_u}\left( \left[ \nabla \ell\left(\x_t\right) \right]_j-\left[ \nabla f  \left(  \widetilde{\x}_t^j, \bar{\y}(\widetilde{\x}_t^j) \right) \right]_j \right)^2+2\sum_{j=1}^{d_u}\left( \left[ \nabla f  \left(  \widetilde{\x}_t^j, \bar{\y}(\widetilde{\x}_t^j) \right) \right]_j-\left[ h_{t}^{f,j} \right]_j \right)^2.
\end{align}

First, we bound the first term on the right-hand side of Eq.~(\ref{eq:error_l_cd}).
\begin{align}\label{eq:error_l_j_cd}
    &\left( \left[ \nabla \ell\left(\x_t\right) \right]_j-\left[ \nabla f  \left(  \widetilde{\x}_t^j, \bar{\y}(\widetilde{\x}_t^j) \right) \right]_j \right)^2 \nonumber\\
    &\le \left\| \nabla \ell\left(\x_t\right)-\nabla f  \left(  \widetilde{\x}_t^j, \bar{\y}(\widetilde{\x}_t^j) \right) \right\|^2 \nonumber\\
    &\le 2\left\| \nabla \ell\left(\x_t\right)-\nabla \ell\left(\widetilde{\x}_t^j\right) \right\|^2+2\left\| \nabla \ell\left(\widetilde{\x}_t^j\right)-\nabla f  \left(  \widetilde{\x}_t^j, \bar{\y}(\widetilde{\x}_t^j) \right) \right\|^2 \nonumber\\
    &\stackrel{(a)}{\le} 2L_l^2\left\| \x_t-\widetilde{\x}_t^j \right\|^2+2L_f^2\left\| \y^*(\widetilde{\x}_t^j)-\bar{\y}(\widetilde{\x}_t^j) \right\|^2 \nonumber\\
    &\stackrel{(b)}{\le} 2L_l^2\left\| \x_t-\widetilde{\x}_t^j \right\|^2+2L_f^2\delta_t^2 \nonumber\\
    &\le 2L_l^2\alpha_{t-1}^2\left\| \mathbf{d}_{t-1}^f \right\|^2+2L_f^2\delta_t^2,
\end{align}
where (a) utilizes the Lipschitzness of $\nabla \ell\left(\x\right)$ and $\nabla f\left(\x, \y\right)$ (see Lemma~\ref{lemma_Lip} and Assumption~\ref{asmp_up}). (b) is due to Assumption~\ref{asmp_y}.

Next, we bound the second term on the right-hand side of Eq.~(\ref{eq:error_l_cd}).
\begin{align}\label{eq:error_f_j_cd}
    &\left( \left[ \nabla f  \left(  \widetilde{\x}_t^j, \bar{\y}(\widetilde{\x}_t^j) \right) \right]_j-\left[ h_{t}^{f,j} \right]_j \right)^2 \nonumber\\
    &\stackrel{(a)}{=}\left(\left[ \nabla_{\x} f  \left(  \widetilde{\x}_t^j, \bar{\y}(\widetilde{\x}_t^j) \right) \right]_j + \left[ \left[\nabla _{\x}\bar{\y}\left( \widetilde{\x}_t^j \right)\right]^\top\nabla_{\y} f\left( \widetilde{\x}_t^j, \bar{\y}\left(\widetilde{\x}_t^j \right) \right) \right]_j\right. \nonumber\\
    &\qquad \left.-\left(\left[ \nabla_{\x} f  \left(  \widetilde{\x}_t^j, \bar{\y}(\widetilde{\x}_t^j) \right) \right]_j + \left[ \left[\widehat{\nabla} _{\x}\bar{\y}\left( \widetilde{\x}_t^j \right)\right]^\top\nabla_{\y} f\left( \widetilde{\x}_t^j, \bar{\y}\left(\widetilde{\x}_t^j \right) \right) \right]_j\right)\right)^2 \nonumber\\
    &\le\left\| \nabla _{\x}\bar{\y}\left( \widetilde{\x}_t^j \right)- \widehat{\nabla} _{\x}\bar{\y}\left( \widetilde{\x}_t^j \right)\right\|^2\left\| \nabla_{\y} f\left( \widetilde{\x}_t^j, \bar{\y}\left(\widetilde{\x}_t^j \right) \right) \right\|^2 \nonumber\\
    &\stackrel{(b)}{\le }B_{f_y}^2\left\| \nabla _{\x}\bar{\y}\left( \widetilde{\x}_t^j \right)- \widehat{\nabla} _{\x}\bar{\y}\left( \widetilde{\x}_t^j \right)\right\|^2 \nonumber\\
    &=B_{f_y}^2\sum_{d=1}^{d_l}\left\| \nabla _{\x}\bar{\y}_d\left( \widetilde{\x}_t^j \right)- \widehat{\nabla} _{\x}\bar{\y}_d\left( \widetilde{\x}_t^j \right)\right\|^2 \nonumber\\
    &\stackrel{(c)}{\le }B_{f_y}^2d_l\frac{L_y^2}{4}\left\| \x_{t-1}-\widetilde{\x}_t^j \right\|^2 \nonumber\\
    &=B_{f_y}^2d_l\frac{L_y^2}{4}\left( x_{t-1}^j-x_t^j \right)^2,
\end{align}
where (a) uses the definitions of $h_{t}^{f,j}$ and $\nabla f\left( \widetilde{\x}_t^j,\bar{\y}\left( \widetilde{\x}_t^j \right) \right)$. (b) is due to Assumption~\ref{asmp_up}. (c) utilizes Lemma~\ref{lemma_zero_cd}.

Substituting Eq.~(\ref{eq:error_l_j_cd}) and Eq.~(\ref{eq:error_f_j_cd}) into Eq.~(\ref{eq:error_l_cd}), we have
\begin{align*}
    &\left\|\nabla \ell\left(\x_t\right)-\mathbf{d}_{t}^f\right\|^2 \\
    &\le 4d_uL_l^2\alpha_{t-1}^2\left\| \mathbf{d}_{t-1}^f \right\|^2+4d_uL_f^2\delta_t^2 + B_{f_y}^2d_l\frac{L_y^2}{2}\left\| \x_{t-1}-\x_t \right\|^2 \\
    &\le 4d_uL_l^2\alpha_{t-1}^2\left\| \mathbf{d}_{t-1}^f \right\|^2+B_{f_y}^2d_l\frac{L_y^2}{2}\alpha_{t-1}^2\left\| \mathbf{d}_{t-1}^f \right\|^2+4d_uL_f^2\delta_t^2.
\end{align*}

Thus, the proof is complete.
\end{proof}

\subsection{Descent in the Potential Function}
We define the potential function $W_t$ as follows:
\begin{equation*}
    W_t=\ell\left( \x_t \right)+K_l\left\|\nabla \ell\left(\x_t\right)-\mathbf{d}_{t}^f\right\|^2.
\end{equation*}

\begin{lemma}\label{lemma_W_cd}
     Under Assumptions~\ref{asmp_up}--\ref{asmp_y}, choose $\alpha_t$ and $K_l$ such that 
     \begin{align*}
         \alpha_t \leq \min &\left\{\frac{1}{2L_l},\frac{1}{32d_uL_l^2K_l}, \frac{1}{4B_{f_y}^2L_y^2d_lK_l} \right\},\quad K_l\ge \frac{\alpha_t}{2}.
    \end{align*}  
    Then, the iterates generated by Algorithm~\ref{alg_cd} satisfies: for all $t\in \left\{ 0,1,\ldots,T-1  \right\}$,
     \begin{align*}
        W_{t+1}-W_t\le -\frac{\alpha_t}{2}\left\|\nabla \ell\left(\x_t\right)\right\|^2+4d_uL_f^2\delta_{t+1}^2K_l.
     \end{align*}
\end{lemma}

\begin{proof}
From Lemma~\ref{apx:lemma_l_cd}, we have 
\begin{align}\label{eq:potential_l_cd}
    &\ell\left(\x_{t+1}\right)-\ell\left(\x_t\right) \le -\frac{\alpha_t}{2}\left\|\nabla \ell\left(\x_t\right)\right\|^2-\frac{\alpha_t}{4}\left\|\mathbf{d}_{t}^f\right\|^2+\frac{\alpha_t}{2}\left\|\nabla \ell\left(\x_t\right)-\mathbf{d}_{t}^f\right\|^2.
\end{align}

Based on Lemma~\ref{apx:lemma_h_cd}, we have 
\begin{align}\label{eq:potential_h_cd}
    &\left\|\nabla \ell\left(\x_{t+1}\right)-\mathbf{d}_{t+1}^f\right\|^2-\left\|\nabla \ell\left(\x_{t}\right)-\mathbf{d}_{t}^f\right\|^2 \nonumber\\
    &\le -\left\|\nabla \ell\left(\x_{t}\right)-\mathbf{d}_{t}^f\right\|^2+4d_uL_l^2\alpha_{t}^2\left\| \mathbf{d}_{t}^f \right\|^2+B_{f_y}^2d_l\frac{L_y^2}{2}\alpha_{t}^2\left\| \mathbf{d}_{t}^f \right\|^2+4d_uL_f^2\delta_{t+1}^2.
\end{align}

Combining Eq. (\ref{eq:potential_l_cd}) and (\ref{eq:potential_h_cd}), we get
\begin{align*}
    & W_{t+1}-W_t\\
    &\le -\frac{\alpha_t}{2}\left\|\nabla \ell\left(\x_t\right)\right\|^2+C_h\left\|\mathbf{d}_{t}^f\right\|^2 +C_l \left\|\nabla \ell\left(\x_{t}\right)-\mathbf{d}_{t}^f\right\|^2+4d_uL_f^2\delta_{t+1}^2K_l,
\end{align*}
where
\begin{align*}
    C_{h}=& -\frac{\alpha_t}{4}+4d_uL_l^2\alpha_{t}^2K_l+B_{f_y}^2d_l\frac{L_y^2}{2}\alpha_{t}^2K_l,\\
    C_{l}=& \frac{\alpha_t}{2}-K_l.
\end{align*}

To ensure $C_{h} \le 0$, we have
\begin{align*}
    C_{h}=& -\frac{\alpha_t}{4}+4d_uL_l^2\alpha_{t}^2K_l+B_{f_y}^2d_l\frac{L_y^2}{2}\alpha_{t}^2K_l\\
    &\stackrel{(a)}{\leq} -\frac{\alpha_t}{4}+\frac{\alpha_t}{8}+\frac{\alpha_t}{8}=0,
\end{align*}
where (a) uses $\alpha_t \leq \min \left\{ \frac{1}{32d_uL_l^2K_l}, \frac{1}{4B_{f_y}^2L_y^2d_lK_l} \right\}$.

To ensure $C_{l} \le 0$, we choose $K_l\ge \frac{\alpha_t}{2}$.

Towards this end, the lemma is proved.
\end{proof}

\subsection{Proof of Theorem~\ref{thm_cd}}

\begin{theorem}[Deterministic Setting]\label{apx:thm_cd}
    Under Assumptions~\ref{asmp_up}--\ref{asmp_y}, choose constant step-sizes $0<\alpha_t=\alpha\le\bar{\alpha}$ for all $t \in \{0,1, \ldots, T-1\}$.
    Moreover, choose $\bar{\alpha}$ such that 
    \begin{align*}
        \bar{\alpha} \leq \min &\left\{\frac{1}{2L_l}, \sqrt{\frac{2}{32d_uL_l^2}}, \sqrt{\frac{2}{4B_{f_y}^2L_y^2d_l}} \right\}.
    \end{align*}    
    Then, the iterates generated by \algcd~satisfy:
    \begin{align*}
        &\frac{1}{T}  \sum_{t=0}^{T-1} \left\|\nabla \ell\left(\x_t\right)\right\|^2 = \mathcal{O}\left( \frac{ \Delta_0 }{T\alpha} \right)+\mathcal{O}\left( d_u\frac{1}{T}  \sum_{t=0}^{T-1}\delta_{t+1}^2 \right),
    \end{align*}
   where $\Delta_0  =  \left( \ell\left( \x_0 \right)-\ell^* \right)+\|\nabla \ell\left( \x_0 \right)-\sum_{j=1}^{d_u}[h_0^{f, j}]_j \e_j\|^2$.
\end{theorem}

\begin{proof}
Let $K_l=\frac{\alpha_t}{2}$, and choose $\alpha_t$ as a constant stepsize $\alpha_t=\alpha$.
Summing the result in Lemma~\ref{lemma_W_cd} from $t=0$ to $T-1$, and then dividing by $T$ on both sides, we get
\begin{align*}
    &\frac{W_T-W_0}{T} \leq -\frac{\alpha}{2T}  \sum_{t=0}^{T-1} \left\|\nabla \ell\left(\x_t\right)\right\|^2 +2d_uL_f^2\alpha\frac{1}{T}  \sum_{t=0}^{T-1}\delta_{t+1}^2.
\end{align*}

Rearranging the terms and multiplying by $2/\alpha$ on both sides, we have
\begin{align*}
    &\frac{1}{T}  \sum_{t=0}^{T-1} \left\|\nabla \ell\left(\x_t\right)\right\|^2 \leq \frac{2\left( W_0-\ell^* \right)}{\alpha T}+4d_uL_f^2\frac{1}{T}  \sum_{t=0}^{T-1}\delta_{t+1}^2,
\end{align*}
where $W_0=\ell\left( \x_0 \right)+K_l\left\|\nabla \ell\left( \x_0 \right)-\mathbf{d}_{0}^f\right\|^2$.

Therefore,
\begin{align*}
    &\frac{1}{T}  \sum_{t=0}^{T-1} \left\|\nabla \ell\left(\x_t\right)\right\|^2 \\
    &= \mathcal{O}\left( \frac{ \left( \ell\left( \x_0 \right)-\ell^* \right)+\left\|\nabla \ell\left( \x_0 \right)-\mathbf{d}_{0}^f\right\|^2 }{\alpha T} \right)+\mathcal{O}\left( \frac{ d_u}{T}  \sum_{t=0}^{T-1}\delta_{t+1}^2 \right).
\end{align*}

The proof of the theorem is completed.
\end{proof}


\section{Theoretical Analysis for \algcd~in Stochastic Setting} \label{apdx: Proof_Stochastic_CD}

Define the coordinate-wise estimator vector
\begin{equation*}
    \widehat{\mathbf{d}}_t^f \triangleq \sum_{j=1}^{d_u}\left[\widehat{h}_t^{f, j}\right]_j \e_j\in \mathbb{R}^{d_u}.
\end{equation*}

The update of the UL parameter can be written as
\begin{equation} \label{eq:x_update_sto}
    \x_{t+1}=\x_{t}-\alpha_{t}\widehat{\mathbf{d}}_t^f.
\end{equation}

\subsection{Descent in the Upper-Level Objective Function}

\begin{lemma}\label{apx:lemma_l_cd_sto}
     Under Assumptions~\ref{asmp_up}--\ref{asmp_sg} and choosing $\alpha_t\le \frac{1}{4L_l}$, the following inequality holds for successive iterations of Algorithm~\ref{alg_cd}:
     \begin{align*}
        &\mathbb{E}\left[ \ell\left(\x_{t+1}\right)-\ell\left(\x_t\right) \right] \\
        &\le -\frac{\alpha_t}{2}\mathbb{E}\left[ \left\|\nabla \ell\left(\x_t\right)\right\|^2 \right]-\frac{\alpha_t}{4}\mathbb{E}\left[ \left\|\mathbf{d}_{t}^f\right\|^2 \right]+\frac{\alpha_t}{2}\mathbb{E}\left[ \left\|\nabla \ell\left(\x_t\right)-\mathbf{d}_{t}^f\right\|^2 \right]+\alpha_t^2 L_l d_u\sigma_h^2,
     \end{align*}
     for all $t\in \left\{ 0,1,\ldots,T-1  \right\}$, where the expectation is taken over the stochasticity of the algorithm.
\end{lemma}

\begin{proof}
We have
\begin{align}\label{eq:lemma_l_cd_sto}
    &\mathbb{E}\left[ \ell\left(\x_{t+1}\right)-\ell\left(\x_t\right) \right] \nonumber\\
    &\stackrel{(a)}{\leq} \mathbb{E}\left[ \left\langle\nabla \ell\left(\x_t\right), \x_{t+1}-\x_t\right\rangle+\frac{L_l}{2}\left\|\x_{t+1}-\x_t\right\|^2 \right]\nonumber\\
    &\stackrel{(b)}{=}\mathbb{E}\left[ -\alpha_t\left\langle\nabla \ell\left(\x_t\right), \widehat{\mathbf{d}}_{t}^f\right\rangle+\frac{\alpha_t^2 L_l}{2} \left\|\widehat{\mathbf{d}}_{t}^f\right\|^2 \right]\nonumber\\
    &\stackrel{(c)}{=}\mathbb{E}\left[ -\alpha_t\left\langle\nabla \ell\left(\x_t\right), \mathbf{d}_{t}^f\right\rangle+\frac{\alpha_t^2 L_l}{2} \left\|\widehat{\mathbf{d}}_{t}^f\right\|^2 \right]\nonumber\\
    &\stackrel{(d)}{=} \mathbb{E}\left[ -\frac{\alpha_t}{2}\left\|\nabla \ell\left(\x_t\right)\right\|^2-\frac{\alpha_t}{2}\left\|\mathbf{d}_{t}^f\right\|^2+\frac{\alpha_t}{2}\left\|\nabla \ell\left(\x_t\right)-\mathbf{d}_{t}^f\right\|^2+\frac{\alpha_t^2 L_l}{2} \left\|\widehat{\mathbf{d}}_{t}^f\right\|^2 \right] \nonumber \\
    &\le \mathbb{E}\left[ -\frac{\alpha_t}{2}\left\|\nabla \ell\left(\x_t\right)\right\|^2-\frac{\alpha_t}{2}\left\|\mathbf{d}_{t}^f\right\|^2+\frac{\alpha_t}{2}\left\|\nabla \ell\left(\x_t\right)-\mathbf{d}_{t}^f\right\|^2+\alpha_t^2 L_l \left\|\widehat{\mathbf{d}}_{t}^f-\mathbf{d}_{t}^f\right\|^2+\alpha_t^2 L_l \left\|\mathbf{d}_{t}^f\right\|^2 \right] \nonumber \\
    &\stackrel{(e)}{\le} \mathbb{E}\left[ -\frac{\alpha_t}{2}\left\|\nabla \ell\left(\x_t\right)\right\|^2-\frac{\alpha_t}{4}\left\|\mathbf{d}_{t}^f\right\|^2+\frac{\alpha_t}{2}\left\|\nabla \ell\left(\x_t\right)-\mathbf{d}_{t}^f\right\|^2+\alpha_t^2 L_l \left\|\widehat{\mathbf{d}}_{t}^f-\mathbf{d}_{t}^f\right\|^2 \right],
\end{align}
where (a) uses the Lipschitz continuous gradients of $\ell$ (see Lemma~\ref{lemma_Lip}). (b) follows from the update rule in Eq.~(\ref{eq:x_update_sto}). (c) utilizes $\mathbb E[\widehat{\mathbf d}_t^f\mid\mathcal F_t]=\mathbf d_t^f$. (d) is because of $\left\langle x,y \right\rangle=\frac{1}{2}\left\| x \right\|^{2}+\frac{1}{2}\left\| y \right\|^{2}-\frac{1}{2}\left\| x-y \right\|^{2}$. (e) chooses $\alpha_t\le \frac{1}{4L_l}$.

Next, we bound the last term on the right-hand side of Eq.~(\ref{eq:lemma_l_cd_sto}).

Define 
\begin{gather*}
    q_t^j= \frac{\bar{\y}(\x_{t-1})-\bar{\y}(\widetilde{\x}_t^j)}{x_{t-1}^j-x_t^j},\qquad
    \widehat{q}_t^j= \frac{\bar{\y}(\x_{t-1};\mathcal{B}_{t-1}^{y})-\bar{\y}(\widetilde{\x}_t^j;\mathcal{B}_t^{y,j})}{x_{t-1}^j-x_t^j} .
\end{gather*}

For each $j\in \left[ d_u \right]$, we have
\begin{align}\label{eq:error_d_cd_sto}
    &\mathbb{E}\left[ \left( \left[ \widehat{\mathbf{d}}_{t}^f-\mathbf{d}_{t}^f \right]_j \right)^2 \right]\nonumber\\
    & =\mathbb{E}\left[ \left( \left[ \nabla_{\x} f \left( \widetilde{\x}_t^j, \bar{\y}(\widetilde{\x}_t^j);\mathcal{B}_t^{f_x} \right) -\nabla_{\x} f  \left(  \widetilde{\x}_t^j, \bar{\y}(\widetilde{\x}_t^j) \right) \right]_j \right.\right.\nonumber\\
    & \left.\left.+\left[ \left\langle \widehat{q}_t^j,\nabla_{\y} f(\widetilde{\x}_t^j, \bar{\y}(\widetilde{\x}_t^j);\mathcal{B}_t^{f_y,j}) \right\rangle -\left\langle q_t^j,\nabla_{\y} f(\widetilde{\x}_t^j, \bar{\y}(\widetilde{\x}_t^j)) \right\rangle \right]_j\right)^2 \right]\nonumber\\
    & =\mathbb{E}\left[ \left(\left[ \nabla_{\x} f \left( \widetilde{\x}_t^j, \bar{\y}(\widetilde{\x}_t^j);\mathcal{B}_t^{f_x} \right) -\nabla_{\x} f  \left(  \widetilde{\x}_t^j, \bar{\y}(\widetilde{\x}_t^j) \right) \right]_j\right.\right.\nonumber\\
    & +\left[ \left\langle q_t^j,\nabla_{\y} f(\widetilde{\x}_t^j, \bar{\y}(\widetilde{\x}_t^j);\mathcal{B}_t^{f_y,j})-\nabla_{\y} f(\widetilde{\x}_t^j, \bar{\y}(\widetilde{\x}_t^j)) \right\rangle +\left\langle \widehat{q}_t^j-q_t^j,\nabla_{\y} f(\widetilde{\x}_t^j, \bar{\y}(\widetilde{\x}_t^j)) \right\rangle \right.\nonumber\\
    &\left.\left.\left.+\left\langle  \widehat{q}_t^j-q_t^j,\nabla_{\y} f(\widetilde{\x}_t^j, \bar{\y}(\widetilde{\x}_t^j);\mathcal{B}_t^{f_y,j})-\nabla_{\y} f(\widetilde{\x}_t^j, \bar{\y}(\widetilde{\x}_t^j))\right\rangle \right]_j\right)^2 \right]\nonumber\\
    & \le 4\mathbb{E}\left[ \left[ \nabla_{\x} f \left( \widetilde{\x}_t^j, \bar{\y}(\widetilde{\x}_t^j);\mathcal{B}_t^{f_x} \right) -\nabla_{\x} f  \left(  \widetilde{\x}_t^j, \bar{\y}(\widetilde{\x}_t^j) \right) \right]_j^2\right]+4\mathbb{E}\left[ \left| \left\langle \widehat{q}_t^j-q_t^j,\nabla_{\y} f(\widetilde{\x}_t^j, \bar{\y}(\widetilde{\x}_t^j)) \right\rangle \right|^2 \right]\nonumber\\
    & + 4\mathbb{E}\left[ \left| \left\langle q_t^j,\nabla_{\y} f(\widetilde{\x}_t^j, \bar{\y}(\widetilde{\x}_t^j);\mathcal{B}_t^{f_y,j})-\nabla_{\y} f(\widetilde{\x}_t^j, \bar{\y}(\widetilde{\x}_t^j)) \right\rangle \right|^2 \right] \nonumber\\
    &+4\mathbb{E}\left[ \left| \left\langle  \widehat{q}_t^j-q_t^j,\nabla_{\y} f(\widetilde{\x}_t^j, \bar{\y}(\widetilde{\x}_t^j);\mathcal{B}_t^{f_y,j})-\nabla_{\y} f(\widetilde{\x}_t^j, \bar{\y}(\widetilde{\x}_t^j))\right\rangle \right|^2  \right].
\end{align}

Then, we bound each term on the right-hand side of Eq.~(\ref{eq:error_d_cd_sto}).
\begin{align}\label{eq:error_d_cd_sto_term1}
    \mathbb{E}\left[ \left[ \nabla_{\x} f \left( \widetilde{\x}_t^j, \bar{\y}(\widetilde{\x}_t^j);\mathcal{B}_t^{f_x} \right) -\nabla_{\x} f  \left(  \widetilde{\x}_t^j, \bar{\y}(\widetilde{\x}_t^j) \right) \right]_j^2\right]\stackrel{(a)}{\leq}\sigma_{f_{x}}^{2},
\end{align}
where (a) is based on Assumption~\ref{asmp_sg}. 
\begin{align}\label{eq:error_d_cd_sto_term2}
    &\mathbb{E}\left[ \left| \left\langle \widehat{q}_t^j-q_t^j,\nabla_{\y} f(\widetilde{\x}_t^j, \bar{\y}(\widetilde{\x}_t^j)) \right\rangle \right|^2 \right]\nonumber\\
    &\stackrel{(a)}{\leq}\mathbb{E}\left[  \left\| \widehat{q}_t^j-q_t^j \right\|^2\left\| \nabla_{\y} f(\widetilde{\x}_t^j, \bar{\y}(\widetilde{\x}_t^j)) \right\|^2 \right]\nonumber\\
    &\stackrel{(b)}{\leq}B_{f_{y}}^2\sigma_y^2,
\end{align}
where (a) is due to Cauchy–Schwarz inequality. (b) uses Assumption~\ref{asmp_up} and Lemma~\ref{lemma_q} with $\mathcal{B}_{t-1}^{y}=\mathcal{B}_t^{y,j}=\mathcal{B}^{y}$.
\begin{align}\label{eq:error_d_cd_sto_term3}
    &\mathbb{E}\left[ \left| \left\langle q_t^j,\nabla_{\y} f(\widetilde{\x}_t^j, \bar{\y}(\widetilde{\x}_t^j);\mathcal{B}_t^{f_y,j})-\nabla_{\y} f(\widetilde{\x}_t^j, \bar{\y}(\widetilde{\x}_t^j)) \right\rangle \right|^2 \right]\nonumber\\
    &\stackrel{(a)}{\leq}\mathbb{E}\left[ \left\| q_t^j \right\|^2\left\| \nabla_{\y} f(\widetilde{\x}_t^j, \bar{\y}(\widetilde{\x}_t^j);\mathcal{B}_t^{f_y,j})-\nabla_{\y} f(\widetilde{\x}_t^j, \bar{\y}(\widetilde{\x}_t^j)) \right\|^2 \right]
    \stackrel{(b)}{\leq} B_y^2\sigma_{f_{y}}^{2},
\end{align}
where (a) uses Cauchy–Schwarz inequality. (b) is due to Assumption~\ref{asmp_sg} and Lemma~\ref{lemma_q}.
\begin{align}\label{eq:error_d_cd_sto_term4}
    &\mathbb{E}\left[ \left| \left\langle  \widehat{q}_t^j-q_t^j,\nabla_{\y} f(\widetilde{\x}_t^j, \bar{\y}(\widetilde{\x}_t^j);\mathcal{B}_t^{f_y,j})-\nabla_{\y} f(\widetilde{\x}_t^j, \bar{\y}(\widetilde{\x}_t^j))\right\rangle \right|^2\right]\nonumber\\
    &\stackrel{(a)}{\leq}\mathbb{E}\left[ \left\| \widehat{q}_t^j-q_t^j \right\|^2\left\| \nabla_{\y} f(\widetilde{\x}_t^j, \bar{\y}(\widetilde{\x}_t^j);\mathcal{B}_t^{f_y,j})-\nabla_{\y} f(\widetilde{\x}_t^j, \bar{\y}(\widetilde{\x}_t^j)) \right\|^2 \right]\nonumber\\
    &\stackrel{(b)}{\leq}\sigma_y^2\sigma_{f_{y}}^{2},
\end{align}
where (a) is because of Cauchy–Schwarz inequality. (b) uses Assumption~\ref{asmp_sg} and Lemma~\ref{lemma_q} with $\mathcal{B}_{t-1}^{y}=\mathcal{B}_t^{y,j}=\mathcal{B}^{y}$.

Substituting Eq.~(\ref{eq:error_d_cd_sto_term1})--(\ref{eq:error_d_cd_sto_term4}) into Eq.~(\ref{eq:error_d_cd_sto}), we obtain
\begin{align*}
    &\mathbb{E}\left[ \left( \left[ \widehat{\mathbf{d}}_{t}^f-\mathbf{d}_{t}^f \right]_j \right)^2 \right] \le 4\sigma_{f_{x}}^{2}+4B_{f_{y}}^2\sigma_y^2 + 4B_y^2\sigma_{f_{y}}^{2}+4\sigma_y^2\sigma_{f_{y}}^{2}=\sigma_h^2.
\end{align*}

Summing over $j=1,2,\dots,d_u$, we get
\begin{align}\label{eq:error_d_sto}
    \mathbb{E}\left[ \left\|\widehat{\mathbf{d}}_{t}^f-\mathbf{d}_{t}^f\right\|^2 \right]\le d_u\sigma_h^2.
\end{align}

Substituting Eq.~(\ref{eq:error_d_sto}) into Eq.~(\ref{eq:lemma_l_cd_sto}) completes the proof.
\end{proof}

\subsection{Descent in the Gradient Approximation Error}
\begin{lemma}\label{apx:lemma_h_cd_sto}
     Under Assumptions~\ref{asmp_up}--\ref{asmp_sg}, the gradient approximation error in Algorithm~\ref{alg_cd} satisfies the following inequality:
     \begin{align*}
        \mathbb{E}\left[ \left\|\nabla \ell\left(\x_t\right)-\mathbf{d}_{t}^f\right\|^2 \right] 
        &\le 8d_uL_l^2\alpha_{t-1}^2\mathbb{E}\left[ \left\| \mathbf{d}_{t-1}^f \right\|^2 \right]+B_{f_y}^2d_lL_y^2\alpha_{t-1}^2\mathbb{E}\left[ \left\| \mathbf{d}_{t-1}^f \right\|^2 \right]\nonumber\\
        &+8d_uL_l^2\alpha_{t-1}^2d_u\sigma_h^2+B_{f_y}^2d_lL_y^2\alpha_{t-1}^2d_u\sigma_h^2+4d_uL_f^2\delta_t^2,
    \end{align*}
    for all $t\in \left\{ 0,1,\ldots,T-1  \right\}$, where the expectation is taken over the stochasticity of the algorithm.
\end{lemma}

\begin{proof}
Following the proof of Lemma~\ref{apx:lemma_h_cd} and using the update rule $\x_{t+1}=\x_{t}-\alpha_{t}\widehat{\mathbf{d}}_t^f$, we have
\begin{align}\label{eq:error_l_cd_det}
    &\mathbb{E}\left[ \left\|\nabla \ell\left(\x_t\right)-\mathbf{d}_{t}^f\right\|^2 \right] \nonumber\\
    &\le 4d_uL_l^2\alpha_{t-1}^2\mathbb{E}\left[ \left\| \widehat{\mathbf{d}}_{t-1}^f \right\|^2 \right]+B_{f_y}^2d_l\frac{L_y^2}{2}\alpha_{t-1}^2\mathbb{E}\left[ \left\| \widehat{\mathbf{d}}_{t-1}^f \right\|^2 \right]+4d_uL_f^2\delta_t^2\nonumber\\
    &\le 8d_uL_l^2\alpha_{t-1}^2\mathbb{E}\left[ \left\| \widehat{\mathbf{d}}_{t-1}^f-\mathbf{d}_{t-1}^f \right\|^2 \right]+8d_uL_l^2\alpha_{t-1}^2\mathbb{E}\left[ \left\| \mathbf{d}_{t-1}^f \right\|^2 \right]\nonumber\\
    &+B_{f_y}^2d_lL_y^2\alpha_{t-1}^2\mathbb{E}\left[ \left\| \widehat{\mathbf{d}}_{t-1}^f-\mathbf{d}_{t-1}^f \right\|^2 \right]+B_{f_y}^2d_lL_y^2\alpha_{t-1}^2\mathbb{E}\left[ \left\| \mathbf{d}_{t-1}^f \right\|^2 \right]+4d_uL_f^2\delta_t^2.
\end{align}

Substituting Eq.~(\ref{eq:error_d_sto}) into Eq.~(\ref{eq:error_l_cd_det}) completes the proof.
\end{proof}

\subsection{Descent in the Potential Function}
We define the potential function $\widehat{W}_t$ as follows:
\begin{equation*}
    \widehat{W}_t=\ell\left( \x_t \right)+K_l\left\|\nabla \ell\left(\x_t\right)-\mathbf{d}_{t}^f\right\|^2.
\end{equation*}

\begin{lemma}\label{lemma_W_cd_sto}
     Under Assumptions~\ref{asmp_up}--\ref{asmp_sg}, choose $\alpha_t$ and $K_l$ such that 
     \begin{align*}
         \alpha_t \leq \min &\left\{\frac{1}{4L_l},\frac{1}{64d_uL_l^2K_l}, \frac{1}{8B_{f_y}^2L_y^2d_lK_l} \right\}, \quad K_l\ge \frac{\alpha_t}{2}.
    \end{align*}  
    Then, the iterates generated by Algorithm~\ref{alg_cd} satisfies: for all $t\in \left\{ 0,1,\ldots,T-1  \right\}$,
     \begin{align*}
        &\mathbb{E}\left[ \widehat{W}_{t+1}-\widehat{W}_t \right]\\
        &\le -\frac{\alpha_t}{2}\mathbb{E}\left[ \left\|\nabla \ell\left(\x_t\right)\right\|^2 \right]+\alpha_t^2 L_l d_u\sigma_h^2+8d_uL_l^2\alpha_{t}^2d_u\sigma_h^2K_l+B_{f_y}^2d_lL_y^2\alpha_{t}^2d_u\sigma_h^2K_l+4d_uL_f^2\delta_{t+1}^2K_l,
     \end{align*}
    where the expectation is taken over the stochasticity of the algorithm.
\end{lemma}

\begin{proof}
From Lemma~\ref{apx:lemma_l_cd_sto}, we have 
\begin{align}\label{eq:potential_l_cd_sto}
    &\mathbb{E}\left[ \ell\left(\x_{t+1}\right)-\ell\left(\x_t\right) \right] \nonumber\\
    &\le -\frac{\alpha_t}{2}\mathbb{E}\left[ \left\|\nabla \ell\left(\x_t\right)\right\|^2 \right]-\frac{\alpha_t}{4}\mathbb{E}\left[ \left\|\mathbf{d}_{t}^f\right\|^2 \right]+\frac{\alpha_t}{2}\mathbb{E}\left[ \left\|\nabla \ell\left(\x_t\right)-\mathbf{d}_{t}^f\right\|^2 \right]+\alpha_t^2 L_l d_u\sigma_h^2.
\end{align}

Based on Lemma~\ref{apx:lemma_h_cd_sto}, we have 
\begin{align}\label{eq:potential_h_cd_sto}
    &\mathbb{E}\left[ \left\|\nabla \ell\left(\x_{t+1}\right)-\mathbf{d}_{t+1}^f\right\|^2-\left\|\nabla \ell\left(\x_t\right)-\mathbf{d}_{t}^f\right\|^2 \right] \nonumber\\
    &\le -\mathbb{E}\left[ \left\|\nabla \ell\left(\x_t\right)-\mathbf{d}_{t}^f\right\|^2 \right] +8d_uL_l^2\alpha_{t}^2\mathbb{E}\left[ \left\| \mathbf{d}_{t}^f \right\|^2 \right]+B_{f_y}^2d_lL_y^2\alpha_{t}^2\mathbb{E}\left[ \left\| \mathbf{d}_{t}^f \right\|^2 \right]\nonumber\\
    &+8d_uL_l^2\alpha_{t}^2d_u\sigma_h^2+B_{f_y}^2d_lL_y^2\alpha_{t}^2d_u\sigma_h^2+4d_uL_f^2\delta_{t+1}^2.
\end{align}

Combining Eq. (\ref{eq:potential_l_cd_sto}) and (\ref{eq:potential_h_cd_sto}), we get
\begin{align*}
    & \mathbb{E}\left[ \widehat{W}_{t+1}-\widehat{W}_t \right]\\
    &\le -\frac{\alpha_t}{2}\mathbb{E}\left[ \left\|\nabla \ell\left(\x_t\right)\right\|^2 \right]+C_h\mathbb{E}\left[ \left\|\mathbf{d}_{t}^f\right\|^2 \right] +C_l \mathbb{E}\left[ \left\|\nabla \ell\left(\x_t\right)-\mathbf{d}_{t}^f\right\|^2 \right]\\
    &+\alpha_t^2 L_l d_u\sigma_h^2+8d_uL_l^2\alpha_{t}^2d_u\sigma_h^2K_l+B_{f_y}^2d_lL_y^2\alpha_{t}^2d_u\sigma_h^2K_l+4d_uL_f^2\delta_{t+1}^2K_l,
\end{align*}
where
\begin{align*}
    C_{h}=& -\frac{\alpha_t}{4}+8d_uL_l^2\alpha_{t}^2K_l+B_{f_y}^2d_lL_y^2\alpha_{t}^2K_l,\\
    C_{l}=& \frac{\alpha_t}{2}-K_l.
\end{align*}

To ensure $C_{h} \le 0$, we have
\begin{align*}
    C_{h}=& -\frac{\alpha_t}{4}+8d_uL_l^2\alpha_{t}^2K_l+B_{f_y}^2d_lL_y^2\alpha_{t}^2K_l\\
    &\stackrel{(a)}{\leq} -\frac{\alpha_t}{4}+\frac{\alpha_t}{8}+\frac{\alpha_t}{8}=0,
\end{align*}
where (a) uses $\alpha_t \leq \min \left\{ \frac{1}{64d_uL_l^2K_l}, \frac{1}{8B_{f_y}^2L_y^2d_lK_l} \right\}$.

To ensure $C_{l} \le 0$, we choose $K_l\ge \frac{\alpha_t}{2}$.

Towards this end, the lemma is proved.
\end{proof}

\subsection{Proof of Theorem~\ref{thm_cd_sto}}

\begin{theorem}[Stochastic Setting]\label{apx:thm_cd_sto}
    Under Assumptions~\ref{asmp_up}--\ref{asmp_sg}, choose constant step-sizes $0< \alpha_t=\alpha\le\bar{\alpha}$ for all $t \in \{0,1, \ldots, T-1\}$.
    Moreover, choose $\bar{\alpha}$ such that 
    \begin{align*}
        \bar{\alpha} \leq \min &\left\{\frac{1}{4L_l}, \sqrt{\frac{2}{64d_uL_l^2}}, \sqrt{\frac{2}{8B_{f_y}^2L_y^2d_l}} \right\}.
    \end{align*}
    Then, the iterates generated by \algcd~satisfy:
    \begin{align*}
        &\frac{1}{T}  \sum_{t=0}^{T-1} \mathbb{E}\left[ \left\|\nabla \ell\left(\x_t\right)\right\|^2 \right] = \mathcal{O}\left( \frac{ \Delta_0 }{T\alpha} \right)+\mathcal{O}\left( d_u \frac{1}{T}  \sum_{t=0}^{T-1}\delta_{t+1}^2+\alpha d_u\left(\sigma_{f_{x}}^{2}+\sigma_y^2 + \sigma_{f_{y}}^{2}+\sigma_y^2\sigma_{f_{y}}^{2} \right) \right),
    \end{align*}
   where $\Delta_0  =  \left( \ell\left( \x_0 \right)-\ell^* \right)+\|\nabla \ell\left( \x_0 \right)-\sum_{j=1}^{d_u}[h_0^{f, j}]_j \e_j\|^2$.
\end{theorem}

\begin{proof}
Let $K_l=\frac{\alpha_t}{2}$, and choose $\alpha_t$ as a constant stepsize $\alpha_t=\alpha$.
Summing the result in Lemma~\ref{lemma_W_cd_sto} from $t=0$ to $T-1$, and then dividing by $T$ on both sides, we get
\begin{align*}
    \frac{\widehat{W}_T-\widehat{W}_0}{T} &\leq -\frac{\alpha}{2T}  \sum_{t=0}^{T-1} \mathbb{E}\left[ \left\|\nabla \ell\left(\x_t\right)\right\|^2 \right] \\
    &+\alpha^2 L_l d_u\sigma_h^2+4d_uL_l^2\alpha^3d_u\sigma_h^2+\frac{1}{2}B_{f_y}^2d_lL_y^2\alpha^3d_u\sigma_h^2+2d_uL_f^2\alpha\frac{1}{T}\sum_{t=0}^{T-1}\delta_{t+1}^2.
\end{align*}

Rearranging the terms and multiplying by $2/\alpha$ on both sides, we have
\begin{align*}
    \frac{1}{T}  \sum_{t=0}^{T-1} \mathbb{E}\left[ \left\|\nabla \ell\left(\x_t\right)\right\|^2 \right] &\leq \frac{2\left( \widehat{W}_0-\ell^* \right)}{\alpha T}\\
    &+2\alpha L_l d_u\sigma_h^2+8d_uL_l^2\alpha^2d_u\sigma_h^2+B_{f_y}^2d_lL_y^2\alpha^2d_u\sigma_h^2+4d_uL_f^2\frac{1}{T}\sum_{t=0}^{T-1}\delta_{t+1}^2,
\end{align*}
where $\widehat{W}_0=\ell\left( \x_0 \right)+K_l\left\|\nabla \ell\left(\x_0\right)-\mathbf{d}_{0}^f\right\|^2$.

Therefore,
\begin{align*}
    &\frac{1}{T}  \sum_{t=0}^{T-1} \mathbb{E}\left[ \left\|\nabla \ell\left(\x_t\right)\right\|^2 \right] \\
    &= \mathcal{O}\left( \frac{ \left( \ell\left( \x_0 \right)-\ell^* \right)+\left\|\nabla \ell\left( \x_0 \right)-\mathbf{d}_{0}^f\right\|^2 }{\alpha T} \right)+\mathcal{O}\left( d_u \frac{1}{T}  \sum_{t=0}^{T-1}\delta_{t+1}^2+\alpha d_u\sigma_h^2 \right).
\end{align*}

The proof of the theorem is completed.
\end{proof}

\section{Broader Impacts} \label{sec: Impacts}

The work presented in this paper aims to advance the field of machine learning by developing algorithms for lower-level agnostic bilevel optimization.
Our work has many potential positive social impacts.
For example, by enabling upper-level learners to operate when the lower-level response is produced by a black-box solver, simulator, adversary, or learned predictor, our \algcd~framework supports a broad range of applications, including adversarial training and deep hyper-representation learning. 
Since this work is foundational and not tied to a specific application domain, we do not anticipate direct negative societal impacts.



\end{document}